\documentclass[10.5pt,compsoc]{tst}
\usepackage{graphicx}
\usepackage{footmisc}
\usepackage{subfigure}
\usepackage{url}
\usepackage{multirow}
\usepackage[noadjust]{cite}
\usepackage{amsmath,amsthm}
\usepackage{amssymb,amsfonts}
\usepackage{booktabs}
\usepackage{color}
\usepackage{ccaption}
\usepackage{booktabs}
\usepackage{float}
\usepackage{fancyhdr}
\usepackage{caption}
\usepackage{xcolor,stfloats}
\usepackage{comment}
\graphicspath{{figures/}}
\usepackage{cuted}  
\usepackage{captionhack}
\usepackage{epstopdf}
\usepackage{amsmath}
\usepackage{bm}
\usepackage{graphicx}
\usepackage[T1]{fontenc} 
\usepackage{xcolor} 
\usepackage{colortbl} 
\usepackage{graphicx} 
\usepackage{float} 
\usepackage{subfigure}
\usepackage{subcaption}
\usepackage{cleveref}  
\headevenname{\normalsize{\textbf{\emph{Tsinghua Science and Technology, xxxxx}} 20xx, 2x(x): xxx--xxx}}%
\headoddname{{\sf Chun Xu et al.:}\quad {\textbf{\emph{Enhancing Aspect-based Sentiment Analysis in Tourism Using Large Language Models and Positional Information}}}}%

\newtheoremstyle{mystyle}{0pt}{0pt}{\normalfont}{1em}{\bf}{}{1em}{}
\theoremstyle{mystyle}
\renewcommand\figurename{Fig.~}

\newcommand{\nop}[1]{}
\makeatletter
\renewcommand{\@biblabel}[1]{[#1]\hfill}
\makeatother
\begin{document}
\thispagestyle{empty}
\hyphenpenalty=50000
\makeatletter
\newcommand\mysmall{\@setfontsize\mysmall{7}{9.5}}
\newenvironment{tablehere}
  {\def\@captype{table}}
  {}
\newenvironment{figurehere}
  {\def\@captype{figure}}
  {}
\thispagestyle{plain}%
\thispagestyle{empty}%
\let\temp\footnote
\renewcommand \footnote[1]{\temp{\normalsize #1}}
{}
\vspace*{-40pt}
\noindent{\normalsize\textbf{\scalebox{0.885}[1.0]{\makebox[5.9cm][s]
{TSINGHUA\, SCIENCE\, AND\, TECHNOLOGY}}}}
\vskip .2mm
{\normalsize
\textbf{
\hspace{-5mm}
\scalebox{1}[1.0]{\makebox[5.6cm][s]{%
I\hfill S\hfill S\hfill N\hfill{\color{white}%
l\hfill l\hfill}1\hfill0\hfill0\hfill7\hfill-\hfill0\hfill2\hfill1\hfill4
\hfill \color{white}{\quad 0\hfill ?\hfill /\hfill ?\hfill ?\quad p\hfill p\hfill  ?\hfill ?\hfill ?\hfill --\hfill ?\hfill ?\hfill ?}\hfill}}}}
\vskip .2mm
{\normalsize
\textbf{
\hspace{-5mm}
\scalebox{1}[1.0]{\makebox[5.6cm][s]{%
DOI:~\hfill~\hfill1\hfill0\hfill.\hfill2\hfill6\hfill5\hfill9\hfill9\hfill/\hfill T\hfill S\hfill T\hfill.\hfill2\hfill0\hfill x\hfill x\hfill.\hfill9\hfill0\hfill1\hfill0\hfill x\hfill x\hfill x}}}}
\vskip .2mm\noindent
{\normalsize\textbf{\scalebox{1}[1.0]{\makebox[5.6cm][s]{%
\color{black}{V\hfill o\hfill l\hfill u\hfill m\hfill%
e\hspace{0.356em}xx,\hspace{0.356em}N\hfill u\hfill%
m\hfill b\hfill e\hfill r\hspace{0.356em}x,\hspace{0.356em}%
x\hfill x\hfill x\hfill x\hfill x\hfill%
x\hfill x\hfill \hspace{0.356em}2\hfill0\hfill x\hfill x}}}}}\\
\begin{strip}
{\center
{\LARGE\textbf{
Enhancing Aspect-based Sentiment Analysis in Tourism Using Large Language Models and Positional Information}}
\vskip 9mm}
{\center {\sf \large
Chun Xu$^{\dagger}$, Mengmeng Wang$^{\dagger}$, Yan Ren, and Shaolin Zhu$^*$
}
\vskip 5mm}
\centering{
\begin{tabular}{p{160mm}}
{\normalsize
\linespread{1.6667} %
\noindent
\bf{Abstract:} {\sf
Aspect-Based Sentiment Analysis (ABSA) in tourism plays a significant role in understanding tourists' evaluations of specific aspects of attractions, which is crucial for driving innovation and development in the tourism industry. However, traditional pipeline models are afflicted by issues such as error propagation and incomplete extraction of sentiment elements. To alleviate this issue, this paper proposes an aspect-based sentiment analysis model, ACOS\_LLM, for Aspect-Category-Opinion-Sentiment Quadruple Extraction (ACOSQE). The model comprises two key stages: auxiliary knowledge generation and ACOSQE. Firstly, Adalora is used to fine-tune large language models for generating high-quality auxiliary knowledge. To enhance model efficiency, Sparsegpt is utilized to compress the fine-tuned model to 50\% sparsity. Subsequently, Positional information and sequence modeling are employed to achieve the ACOSQE task, with auxiliary knowledge and the original text as inputs. Experiments are conducted on both self-created tourism datasets and publicly available datasets, Rest15 and Rest16. Results demonstrate the model's superior performance, with an F1 improvement of 7.49\% compared to other models on the tourism dataset. Additionally, there is an F1 improvement of  0.05\% and 1.06\% on the Rest15 and Rest16 datasets, respectively.}
}
\end{tabular}
}
\vskip 6mm
\vskip -3mm
\small\end{strip}
\thispagestyle{plain}%
\thispagestyle{empty}%
\makeatother
\pagestyle{tstheadings}
\begin{figure}[b]
\vskip -6mm
\begin{tabular}{p{44mm}}
\toprule\\
\end{tabular}
\vskip -4.5mm
\noindent
\setlength{\tabcolsep}{1pt}
\begin{tabular}{p{1.5mm}p{79.5mm}}
$\bullet$& Chun Xu, Mengmeng Wang, and Yan Ren are with the Xinjiang University of Finance and Economics, Urumqi, 830000, China. E-mail: xuchun@xjufe.edu.cn; 17828173236@163.com; yhat@xjufe.edu.cn
\\
$\bullet$& Shaolin Zhu is with the College of Intelligence and Computing, Tianjin University, Tianjin, 300000, China. E-mail: zhushaolin@tju.edu.cn
\\
${\dagger}$& Chun Xu and Mengmeng Wang contribute equally to this paper.
\\
$\ast$& To whom correspondence should be addressed.
\\
&          Manuscript received: year-month-day;
%revised: year-month-day;
accepted: year-month-day

\end{tabular}
\end{figure}\large
\section{Introduction}
\label{s:introduction}
\noindent
With the rapid growth of the tourism industry, individuals increasingly share their travel experiences and emotional feedback through online reviews, which contain valuable insights for planning travel itineraries\textsuperscript{\cite{1}}. 
Traditional sentiment analysis methods typically focus solely on the overall sentiment polarity of sentences, overlooking specific aspects of sentiment\textsuperscript{\cite{2,3,4}}. 
Aspect-Based Sentiment Analysis (ABSA) in tourism  aims to explore tourists' sentiment experiences towards particular aspects of tourism, aiding in understanding user preferences, optimizing products, and fostering the intelligent development of the tourism industry\textsuperscript{\cite{5}}.

\begin{figure*}[htb]
	\centering
	\includegraphics[width=\textwidth]{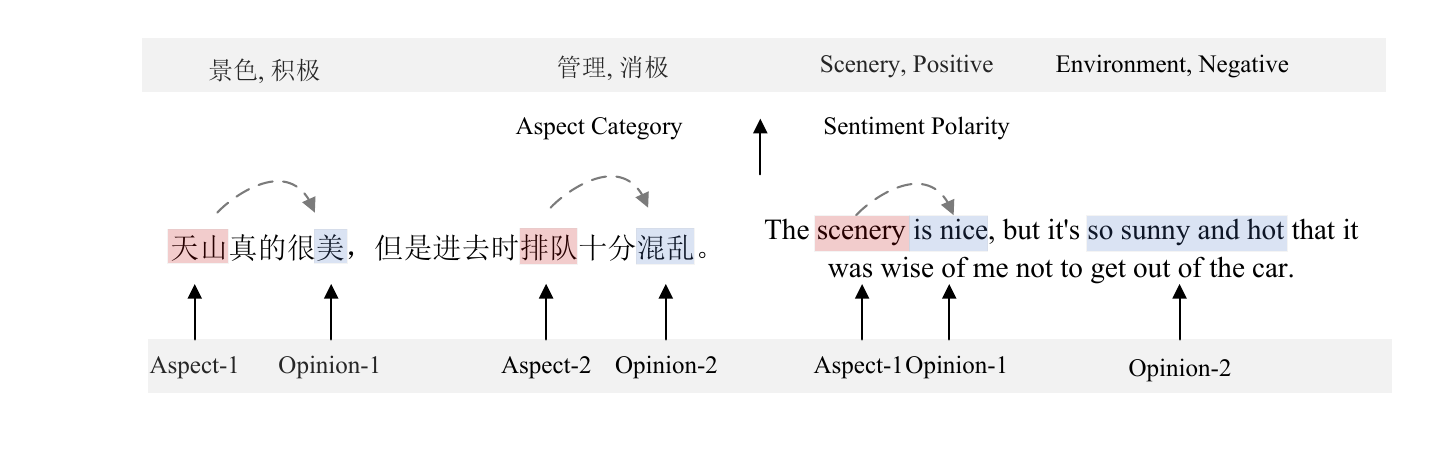}
	\vspace{-12mm}
	\caption{Aspect-category-opinion-sentiment quadruple extraction diagram}
	\vspace{-3mm}
	\label{fig:1}
\end{figure*}

The ABSA task comprises four main subtasks: aspect term extraction, opinion term extraction, aspect polarity classification, and aspect category classification\textsuperscript{\cite{6}}. Currently, researchers primarily focus on individual tasks or joint tasks, including category-sentiment pair extraction, aspect-category-sentiment pair extraction, and aspect-opinion-sentiment pairs extraction\textsuperscript{\cite{7,8,9,10}}. Taking Xinjiang tourism review data as an example, we have completed the Aspect-Category-Opinion-Sentiment Quadruple Extraction (ACOSQE) task to achieve comprehensive sentiment analysis. As illustrated in Figure \ref{fig:1}, our objective is to extract aspect terms and opinion terms from the text and predict their aspect categories and sentiment polarities. For texts that describe emotions without explicitly identifying the subject or object, we assign the aspect term as ``None''. We then continue with extracting sentiment words, and classifying both aspects and sentiments, as illustrated in the second sentence of Figure \ref{fig:1}. This task helps alleviate the issue of missing sentiment quadruples extraction in the tourism domain, providing strong support for more comprehensive tourism sentiment analysis\textsuperscript{\cite{11}}.
\par
In the early phases, researchers employed manually designed features and rules to capture sentiments related to specific aspects in the text, such as sentiment lexicons and bag-of-words models\textsuperscript{\cite{12}}. However, constrained by the quality of features, this approach struggled to handle the complexity of language structures and contextual information\textsuperscript{\cite{13}}. With the advent of statistical natural language processing methods, techniques like support vector machines and naive bayes were utilized by researchers to learn the relationship between features and aspect sentiment. Nonetheless, challenges persisted regarding feature quality and model generalization capability\textsuperscript{\cite{14,15}}. Subsequently, deep learning-based methods emerged as a new paradigm in ABSA. For instance, long short-term memory networks have been employed to enhance the model's capacity to handle complex language and contextual information by modeling text sequences\textsuperscript{\cite{16,17,18,19}}. Graph convolutional networks have also been utilized to enhance the model's understanding and representation of sentiment, as they can comprehensively consider both textual content and relationships between texts. Nevertheless, these models exhibit poor generalization performance on data outside the training scope\textsuperscript{\cite{20,21}}.
\par
In contrast, pre-trained models have been leveraged for ABSA tasks, demonstrating improved generalization to tasks across diverse domains\textsuperscript{\cite{22,23,24,25,26}}. For instance, the Bidirectional Encoder Representations from Transformers (BERT) model learns bidirectional contextual information during pre-training, which enhances the model's ability to understand complex relationships and capture contextual information\textsuperscript{\cite{27}}. Recently, the emergence of Large Language Model (LLM) has opened up new avenues for sentiment analysis, exemplified by the GPT series\textsuperscript{\cite{28,29,30}}. Their ability to learn universal language representations from extensive corpora grants them strong transferability. Consequently, they have been utilized to address the challenge of limited labeling resources in specific domains\textsuperscript{\cite{31,32,33}}. However, despite the potential of LLMs to generate plausible answers in ACOSQE tasks, they still exhibit low accuracy. This is attributed to the stringent nature of precise matching evaluation, whereby the flexibility of LLM-generated outputs often leads to discrepancies with the textual expression.
\par
Thus, we utilize the auxiliary knowledge produced by LLM and the original text as inputs to assist the ACOSQE task. Moreover, issues related to over-parameterization are often encountered by LLMs. To mitigate the challenge of transferring prior knowledge to the task of emotion extraction in tourism, we achieve model compression through weight pruning to eliminate redundant parameters. Sequence models are capable of effectively capturing sequential information and contextual relationships in textual data, and are often used for sentiment and aspect term extraction\textsuperscript{\cite{34,35,36,37}}. We utilize Bi-directional Long Short-Term Memory (BiLSTM) to assist in aspect term and sentiment word extraction. The surrounding text content of sentiment words can influence the true meaning of the sentiment expression. Therefore, we utilize positional information and Bidirectional Gated Recurrent Unit (BiGRU) to help the model understand the contextual environment of sentiment expressions and distinguish the sentiment orientation of different aspects within the same text\textsuperscript{\cite{38,39}}. The main 
contributions are outlined as follows:
\begin{itemize}
	\item To enhance the performance of the tourism sentiment quadruple extraction task, we fine-tuned LLM to generate targeted auxiliary knowledge and analyzed the specific impact of similar types of auxiliary knowledge on model performance. Experimental validation shows that our model excels in extracting tourism sentiment quadruples and also demonstrates good performance in other subtasks of aspect-level sentiment analysis.
	\item We employ neural ordinary differential equations to simulate the discrete processes of stacked neural network components for dynamic adjustments to attention weights. This approach not only enhances the modeling of the relative positional dependencies between context and aspect terms but also helps to mitigate the parameter load associated with stacked attention layers.
\end{itemize}
\section{Related work}
\label{s:Related work}
\noindent
Presently, two primary approaches are employed to address the ACOSQE task: pipeline-based models and end-to-end learning. In pipeline-based models, the ACOSQE task is divided into independent steps, namely aspect extraction, opinion extraction, polarity classification, and category classification, which are processed sequentially, with the output of each step serving as the input for the next step. Zhang et al.\textsuperscript{\cite{40}} introduced a two-stage neural network model that comprises multiple modules. Initially, this model extracts sentiment aspects and opinion terms, identifies their respective categories and polarities. Subsequently, it verifies the corresponding relationships between aspects and opinions. The BiLSTM was utilized to capture contextual information in input sequences by this model. Cai et al.\textsuperscript{\cite{41}} used BERT to obtain context-aware text representations. This method extracts aspect-opinion pairs and models category-emotion classification as a multi-class problem, and then predicts category-emotion based on the extracted aspect-opinion pairs. Pipeline-based aspect-level sentiment analysis methods have the advantages of modularity and strong interpretability, but it suffer from error propagation and difficulty in handling complex relationships between different aspects\textsuperscript{\cite{42,43}}.
\par
Some scholars adopt an end-to-end approach\textsuperscript{\cite{44}}. Xiong et al.\textsuperscript{\cite{45}} treated ACOSQE task as a sequence generation task and captures dependencies among quadruplets through contextual learning, thus providing a more comprehensive understanding of the context. Zhang et al.\textsuperscript{\cite{46}} aimed to enhance the model's adaptability to context by mapping target quadruplets into natural language forms and fine-tuning T5 using input-target pairs. These end-to-end models extract sentiment information directly from raw text, thereby mitigating the accumulation of errors across processing stages, capturing interactions among different text components, and enhancing the model's adaptability to complex relationships. At the same time, this method can capture the interaction between different components in the text, thereby aiding in improving the model's adaptability to complex relationships\textsuperscript{\cite{47,48,49}}.
\par
Introducing pretrained models into low-resource scenarios can alleviate the challenges posed by data scarcity, as these models are trained on vast amounts of text data and acquire general capabilities in language representations and understanding. For example, Atom-7B, built upon the foundation of Llama2\textsuperscript{\cite{50}}, was trained on a large-scale Chinese dataset to enhance its ability to process Chinese text. Additionally, techniques such as positional interpolation and neural tangent kernel are leveraged for adaptive context expansion, enabling the model to consider a broader range of contextual information when processing inputs. Simultaneously, FlashAttention-2 was utilized by Atom-7B to enhance the efficiency of attention mechanisms and optimize model performance on long sequence contexts. Roumeliotis et al.\textsuperscript{\cite{51}} fine-tuned LLM using a large-scale product review dataset to predict product review ratings.
\par
The practical deployment of LLMs face challenges due to their immense scale and high computational costs, prompting increased attention to model compression\textsuperscript{\cite{52}}. Pruning is one of the commonly used model compression methods, which reduces the model's size by removing unimportant connections or nodes. Early methods frequently used weight magnitude as a pruning criterion, removing weights below a certain threshold, but they did not consider the input's impact on the results in this approach. Sun et al.\textsuperscript{\cite{53}} incorporated both weights and input into the calculation of weight importance, comparing weights based on the output to obtain relative importance information. Frantar et al.\textsuperscript{\cite{54}} proposed a post-training pruning method for LLMs, which optimized the retained weights while keeping the mask selection weights unchanged. Experimental results demonstrated that this method maintained model performance and improved generalization ability.
\section{The proposed model}
\noindent
\subsection{Task description}
\noindent
\begin{figure*}[htb]
	\centering
	\includegraphics[width=\textwidth]{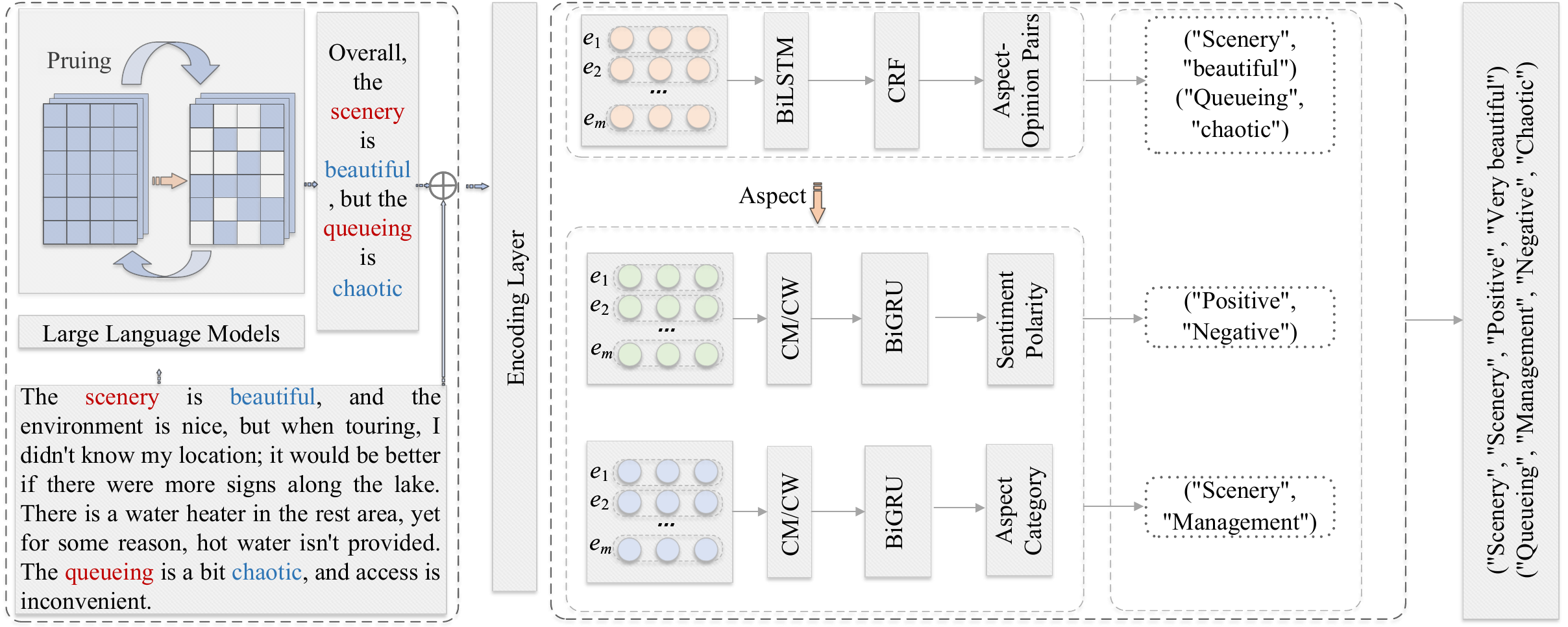}
	\vspace{-5mm}
	\caption{ Model overall structure diagram}
	\vspace{-2mm}
	\label{fig:2}
\end{figure*}
Given a dataset \( D \) comprised of input sequences formatted as sentences \( X = \{W_1, W_2, \dots, W_n\} \), where \( W_n \) represents the \( n \)-th word in sentence $X$, our objective is to extract all sentiment elements $(c, a, o, p)$ from $X$. Here, \( c \) represents the aspect category, \( a \) denotes the aspect term, \( o \) indicates the opinion term, and \( p \) specifies the sentiment polarity. The aspect category \( c \) is restricted to a predefined set \{scenery, service, management, environment, price, attraction, facility, feature\}. Both \( a \) and \( o \) are defined as specific text spans within sentence \( X \), while \( p \) is categorized into one of the following sentiment set \{positive, negative, neutral\}. The model structure, as illustrated in Figure \ref{fig:2}, comprises two key stages: auxiliary knowledge generation and ACOSQE. Initially, the pruned LLM generates auxiliary text, which is then combined with the original input text and encoded using BERT, where \( e_m \) refers to the embedding vector at position \( m \). Following this, aspect-opinion pair extraction and sentiment classification task are jointly trained, where Context Mask (CM) and Context Weight (CW) are used to assist in sentiment and aspect classification, BiLSTM and BiGRU for context modeling, and finally Conditional Random Fields (CRF) for sequence modeling. Finally, the results are parsed into quadruples.
\subsection{The stage of auxiliary knowledge generation}
\noindent
High-quality auxiliary knowledge is introduced to enhance the performance of the travel ABSA task, aiming to alleviate the complexity of data annotation and the lack of public datasets. Specifically, LLMs are utilized for text summarization, which is combined with the original text as input to assist the ACOSQE module. In this paper, Atom-7B is utilized as the auxiliary knowledge generation model, which is based on the transformer architecture and has been optimized, while being trained on a large-scale Chinese dataset. Firstly, the input of each transformer layer is normalized, with the first layer normalization moved before the multi-head self-attention layer and the second layer normalization moved before the fully connected layer. At the same time, the position of residual connections is adjusted after the multi-head self-attention layer and before the fully connected layer. The root mean square normalization function\textsuperscript{\cite{55}} is used in layer normalization, and the specific calculations are shown in equations \eqref{1} and \eqref{2}.
\begin{equation}
	\text{RMS}(\boldsymbol{a})=\sqrt{\frac{1}{n} \sum_{i=1}^n \boldsymbol{a}_i^2}\label{1}
\end{equation}
\begin{equation}
	\overline{\boldsymbol{a}}_i=\frac{\boldsymbol{a}_{\boldsymbol{i}}}{\text{RMS}(\boldsymbol{a})}\label{2}
\end{equation}
Where $ \bm{a} $ is the input vector, $ \bm{a}_i $ is $i$-th element of vector $ \bm{a} $, $n$ is the number of elements in vector $ \bm{a} $, RMS is the square root of the average of the squares of the elements in vector $ \bm{a} $, and $\overline{\boldsymbol{a}}_i$ is the normalized value of $ \bm{a}_i $. Secondly, to improve the non-linear modeling ability of the neural network, SwiGLU\textsuperscript{\cite{56}} is used as the activation function, and the calculation process is shown in the following equation.
\begin{equation}
	\text{FFN}_\text{SwiGLU}=\operatorname{SwiGLU}(\boldsymbol{x}, \boldsymbol{W}, \boldsymbol{V}) \boldsymbol{W}_2\label{3}
\end{equation}
\begin{equation}
	\operatorname{SwiGLU}(\boldsymbol{x}, \boldsymbol{W}, \boldsymbol{V})=\operatorname{Swish}_\beta(\boldsymbol{x} \boldsymbol{W}) \otimes \boldsymbol{x} \boldsymbol{V}\label{4}
\end{equation}
\begin{equation}
	\operatorname{Swish}_\beta(\boldsymbol{x})=\boldsymbol{x} \sigma(\beta \boldsymbol{x})\label{5}
\end{equation}
Where SwiGLU is an activation function, $\text{FFN}_\text{SwiGLU}$ refers to the output of the feedforward network that uses SwiGLU as the activation function, \( \bm{x} \) denotes the input vector, \( \bm{W} \), \( \bm{V} \), and \( \bm{W}_2 \) are defined as the weight matrices, Swish\(_\beta\) is the Swish activation function parameterized by \(\beta\), $\sigma(\bm{x})$ represents the Sigmoid function, 
\par
Subsequently, to alleviate the issues of traditional absolute position encoding, which are prone to length constraints and lack of translational invariance, rotational position embedding\textsuperscript{\cite{57}} is adopted to replace the original absolute position encoding. Long-range dependencies between different positions are captured by adding relative positional information for queries ($\bm{q}$) and keys ($\bm{k}$), as shown in equation \eqref{6}. In the two-dimensional case, the rotary position embedding is represented as equation \eqref{7}.
\begin{equation}
	\tilde{\boldsymbol{q}}_m=f(\boldsymbol{q}, m), \tilde{\boldsymbol{k}}_n=f(\boldsymbol{k}, n)\label{6}
\end{equation}
\begin{equation}
	f(\boldsymbol{q}, m)=\left(\begin{array}{cc}
		\cos m \theta & -\sin m \theta \\
		\sin m \theta & \cos m \theta
	\end{array}\right)\left(\begin{array}{l}
		q_0 \\
		q_1
	\end{array}\right)\label{7}
\end{equation}
Where $\tilde{\boldsymbol{q}}_m$ and $\tilde{\boldsymbol{k}}_n$  are the transformed query and key vectors at positions $m$ and $n$ respectively, $\theta$ is the angle with absolute positional information, $f$ is transformation function, \( q_0\) and \(q_1 \) are components of the vector \( \boldsymbol{q} \),
\par
Finally, to generate results more in line with human preferences, human feedback is utilized as a reward signal for optimizing the model. This involves taking the model-generated responses along with corresponding prompts as input and outputting a scalar score indicating the quality of the generated responses. The loss function for this is represented as equation \eqref{8}.
\begin{equation}
	L=-\log \left(\sigma\left(r_\theta\left(x, y_{c}\right)-r_\theta\left(x, y_{r}\right)\right)-m(r))\right)\label{8}
\end{equation}
Where $L$ represents the loss function, $r_\theta(x, y)$ represents the score obtained by the model when evaluating the prompt $x$ and the output $y$ with model weights $\theta$, $y_c$ is the preferred response selected by the annotator, $y_r$ is the rejected response, and $\sigma$ represents the Sigmoid function. The boundaries of the discrete function $m(r)$ are determined based on different preference scores.
\subsection{The stage of ACOSQE}
\noindent
The stage of ACOSQE consists of two parts: Aspect-Opinion Pairs Extraction (AOE) and Aspect Category Sentiment Classification (ACS). To enhance the feature representation capability of the model, a joint training approach is adopted, with BERT being used as the word embedding model.
\subsubsection{Aspect-opinion pairs extraction}
\begin{figure}[htb]
	\centering
	\includegraphics[width=\linewidth]{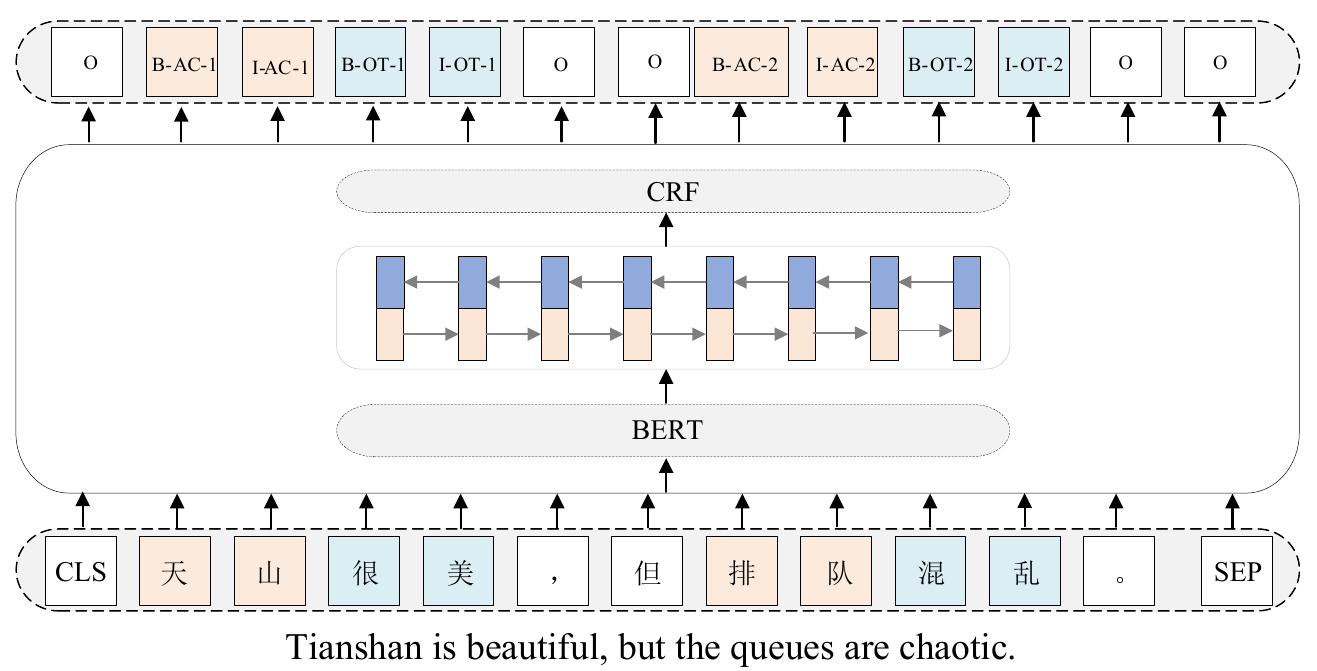}
	\vspace{-6mm}
	\caption{Example of aspect opinion pairs extraction}
	\vspace{-1mm}
	\label{fig:3}
\end{figure}

\noindent
Due to the presence of multiple aspects in the text, along with different sentiment words associated with each, extracting both aspect terms and sentiment words necessitates discerning their respective relationships. Therefore, numerical suffixes are added to entity annotations, as illustrated in Figure \ref{fig:3}, to distinguish different aspect terms. The Begin Inside Outside method is employed for entity tagging, where B-AC denotes the beginning of aspect terms, B-OT denotes the beginning of opinion terms to distinguish entity attributes, I-AC is the inside of an aspect term, I-OT is the inside of an opinion term, and O denotes outside any entity. To establish their corresponding relationship, ``aspect-1'' is designated as B-AC-1, while its corresponding opinion term is designated as B-OT-1. This approach transforms the task of joint entity relation extraction into a sequence labeling problem, effectively sidestepping the need for complex feature engineering.
\par
As shown in Figure \ref{fig:3}, BERT is used for initial text encoding, followed by BiLSTM for capturing contextual information, and CRF for decoding in the AOE task. CRF is commonly used to  enhance sequence prediction accuracy by imposing prior constraints on transition probabilities and label occurrences, ensuring that generated sequences adhere to dependencies between adjacent labels. The evaluation calculation is depicted in equation \eqref{9}, while the probability of the predicted sequence $Y$ is represented by equation \eqref{10}.
\begin{equation}
	S(x, y)=\sum_{i=0}^n \boldsymbol{C}_{y_i, y_{i+1}}+\sum_{i=1}^n \boldsymbol{P}_{i, y_i}	\label{9}
\end{equation}
\begin{equation}
	P(Y \mid X)=\frac{e^{S(x, y)}}{\sum_{Y \in Y_x} e^{S(x, y)}}\label{10}
\end{equation}
Where $S(x,y)$ is scoring function for a given input sequence $x$ and output sequence $y$, $n$ is the length of the sequence $y$, $\bm{C}$ is the transition score matrix, $\bm{P}$ is the output score matrix, $P(Y \mid X)$ is the probability of a specific sequence $Y$ given the input sequence $X$, and $Y_x$ represents all possible sequences.
\subsubsection{Aspect category sentiment classification}
\begin{figure}[htb]
	\centering
	\vspace{1mm}
	\includegraphics[width=\linewidth]{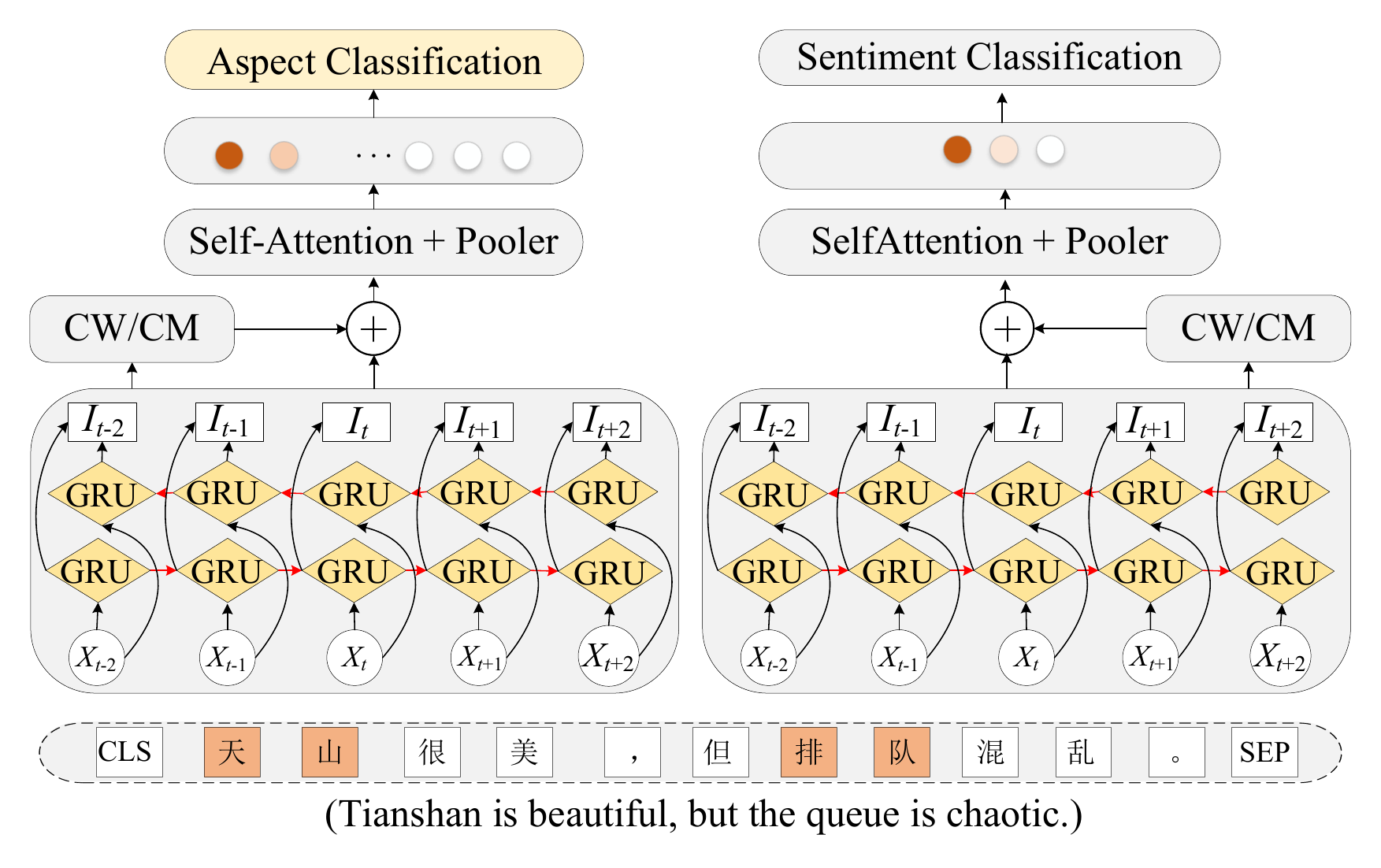}
	\caption {Example of aspect category sentiment classification}
	\label{fig:4}
\end{figure}

\noindent
The aspect term extraction module's results serve as the aspect for prediction in ACS task. ACS takes this input and outputs its corresponding category and sentiment polarity, as illustrated in Figure \ref{fig:4}.
\par
As shown in Figure \ref{fig:4}, the ACS module primarily accomplishes two tasks: sentiment classification and aspect classification. Firstly, to enhance the model's capability of capturing temporal information within sequential data, Gated Recurrent Units (GRU) are employed for modeling temporal sequences after text encoding, where $X$ represents the input, and $I$ denotes the intermediate state. Secondly, context-feature weighting and context-feature mask are utilized to assign varied weights to the context surrounding aspect terms, given the presence of multiple aspects in the text, and the likelihood that each aspect corresponds to different sentiments. This approach allows the model to focus on the context surrounding the entities, thus enhancing contextual information. The calculation of $\mathit{\text{cm}}[i]$ is shown in equation \eqref{11}.
\begin{equation}
	\operatorname{cm}[i]=\left\{\begin{array}{l}
		1, \text { if } \min \left(\left|x_1\right|,\left|x_2\right| )\leq \operatorname{SRD}\right. \\
		0, \text { otherwise }
	\end{array}\right.
	\label{11}
\end{equation}
\vspace{-2mm}
\\
Where cm denotes the mask feature, $x_1$ is $i$ - ent\_pos[0], $x_2$ is $i$ - ent\_pos[-1], ent\_pos[0] is the starting position of the entity, ent\_pos[-1] is the ending position, $\min \left(\left|x_1\right|, \left|x_2\right|\right)$ is minimum distance from the current index $i$ to either the starting or ending position of the entity and SRD represents the semantic relative distance threshold. The calculation of CW is as shown in equation \eqref{12}.
\vspace{-0.7mm}
\begin{equation}
	\operatorname{cw}[i]=\frac{\min \left(\left|x_1\right|,\left|x_2\right|\right)-\operatorname{SRD}+1}{\text{max\_len}}\label{12}
\end{equation}
Where cw represents the weight feature, max\_len represents the maximum possible length of context considered, if $\operatorname{cm}[i]$ is 1, then $\operatorname{cw}[i]$ is also 1.
\par
\begin{figure}[h]
	\centering
	\includegraphics[width=\linewidth]{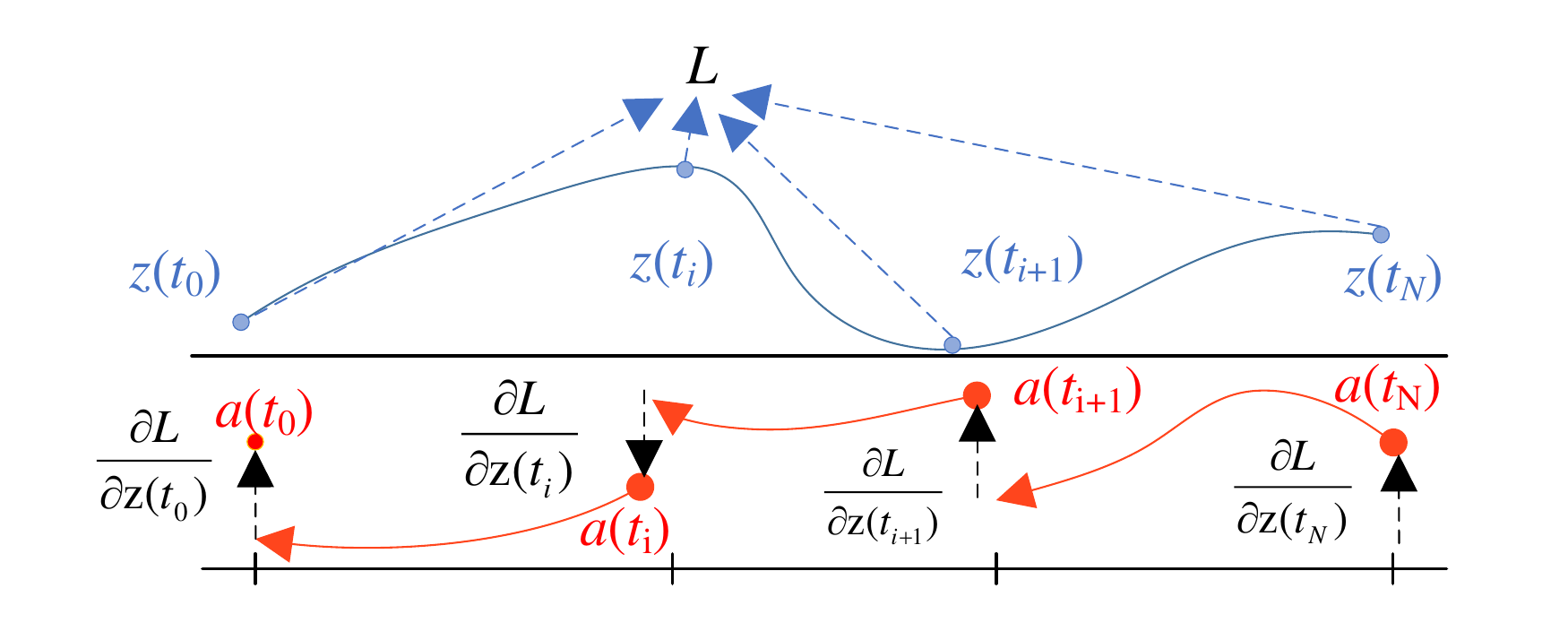}
	\caption{Reverse-mode differentiation of an ODE solution}
	\vspace{-3mm}
	\label{fig:5}
\end{figure}

To further capture the dependency between sentiment aspects and context, after obtaining the initial positional weights, the Neural Ordinary Differential Equations (ODE) module is used to adjust the relative positional attention of the context. The formula for the ODE system is shown in equation \eqref{13} below, using a neural network with learnable parameters to simulate the rate of change of the state, with the hidden state at the final time $t$ used as the positional attention. As shown in Figure \ref{fig:5}, where $L$ is the loss function, where $z$ represents the state trajectory function, $t$ denotes the time, and $a$ is the adjoint state. The Neural ODE illustrates the process of a variable changing over time and through the backpropagation process. During gradient backpropagation, the Neural ODE continuously optimizes the same layer of parameters, which amounts to 0.142 M. We utilize the ``odeint'' function from the ``torchdiffeq'' library to solve the ODE, with the solving method set to dopri5 and a relative tolerance of 0.1.
\begin{equation}
	\frac{d\bm{z}(t)}{dt} = F(\bm{z}(t), t; \bm{\theta})\label{13}
\end{equation}
where \( \bm{z}(t) \) represents the state at time \( t \), and \( F \) is a function parameterized by the network parameters \( \bm{\theta} \).

Next, to address potential instability of local information due to data noise, text features incorporating distance information are merged with globally encoded features from BERT to assist the model in understanding the overall semantics and logical structure of the text. Additionally, the self-attention mechanism is employed to capture the relationships between different elements in the concatenated feature vectors. Finally, head pooling is utilized to extract the hidden state corresponding to the first token position in the input sequence, and sentiment and polarity categories are predicted through softmax.
\subsection{Model optimization}
\subsubsection{Fine-tuning Atom-7B using Adalora}
\begin{figure}[htb]
	\centering
	\vspace{-3mm} 
	\includegraphics[width=\linewidth]{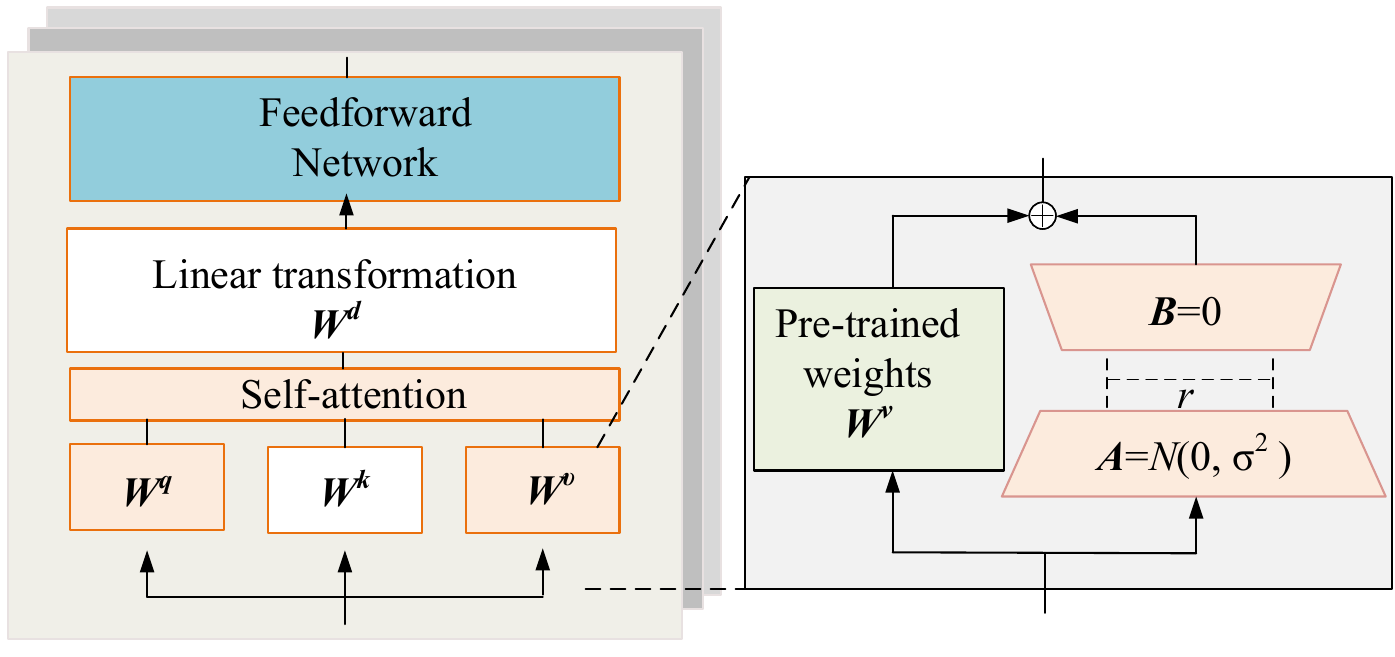}
	\caption{Lora architecture diagram}
	\label{fig:6}
\end{figure}

\noindent
Supervised fine-tuning has been employed to assist LLMs in performing effectively in the target domain\textsuperscript{\cite{58}}. However, due to the vast number of parameters in LLMs, fine-tuning all parameters requires considerable computational power. Reference\textsuperscript{\cite{59}} indicates that language models typically have low intrinsic rank in weight matrices after fine-tuning for specific tasks. Lora introduces the low-rank matrices \( \bm{A} \) and \( \bm{B} \) alongside the original weight matrices to enhance model adaptability with minimal parameter increase, as shown in Figure \ref{fig:6}. In this figure, \( \bm{W}^{\bm{q}} \), \( \bm{W}^{\bm{k}} \), and \( \bm{W}^{\bm{v}} \) denote the matrices for queries, keys, and values, respectively. The matrix \( \bm{W}^{\bm{d}} \) functions as the linear transformation in the feedforward network. The parameters \( \bm{B} \) and \( \bm{A} \) are trainable, and are typically initialized from a normal distribution \( N \) with a mean of zero and a variance of \( \sigma^2 \), $r$ represents the rank. The product is used as trainable parameters by Lora to simulate changes in parameters\textsuperscript{\cite{60}}, as shown in equation \eqref{14}.
\begin{equation}
	\boldsymbol{h}=(\boldsymbol{W}^{(0)}+\Delta\bm{W})\boldsymbol{x}=(\boldsymbol{W}^{(0)}+\boldsymbol{B} \boldsymbol{A})\boldsymbol{x}\label{14}
\end{equation}
Where \(\bm{h}\) is the output after adding Lora, \(\bm{x}\) is the input.
\par
The significant differences in the importance of weight matrices between different modules in the Lora fine-tuning strategy are overlooked. However, Adalora actively adjusts the rank size based on the importance of each weight matrix to downstream tasks during the fine-tuning process. As a result, this method can reduce the number of trainable parameters while maintaining model performance\textsuperscript{\cite{61}}. Initially, let \(\Delta\bm{W} = \bm{P}\Lambda\bm{Q}\), where \(\Lambda\) is a diagonal matrix, and \(\bm{P}\) and \(\bm{Q}\) are orthogonal matrices resulting from the singular value decomposition of the trainable parameter \(\Delta\bm{W}\). Simultaneously, a regularization term is added to the loss function to control the size of the parameters, avoiding numerical instability caused by excessively large parameter values and preventing overfitting of the model on the training data. The regularization term is represented by equation \eqref{15}, and the loss function by equation \eqref{16}. Then, parameters are updated through gradient backpropagation to compute the approximate solution of the weight matrix and its singular value decomposition. Finally, importance scores are calculated using singular values and their singular vectors to prune unimportant singular values.
\begin{equation}
	R(\boldsymbol{P}, \boldsymbol{Q})=\left\|\boldsymbol{P}^T \boldsymbol{P}-\boldsymbol{I}\right\|_F^2+\left\|\boldsymbol{Q}^T \boldsymbol{Q}-\boldsymbol{I}\right\|_F^2\label{15}
\end{equation}
\begin{equation}
	L(\boldsymbol{P}, \boldsymbol{\lambda}, \boldsymbol{Q})=C(\boldsymbol{P}, \boldsymbol{\lambda}, \boldsymbol{Q})+\gamma \sum_{k=1}^n R\left(\boldsymbol{P}_k, \boldsymbol{Q}_k\right)\label{16}
\end{equation}
Where $R(\boldsymbol{P}, \boldsymbol{Q})$ is penalizes the deviation of \(\bm{P}\) and \(\bm{Q}\) from orthonormal matrices, \(\bm{I}\) is the identity matrix. $L$ is loss function, \(C\) is the cost of parameter training, $\boldsymbol{\lambda}$ is diagonal matrix in the decomposition, and \(\gamma\) is the regularization coefficient.
\subsubsection{Model pruning}
\begin{figure}[htb]
	\centering
	\includegraphics[width=1\linewidth]{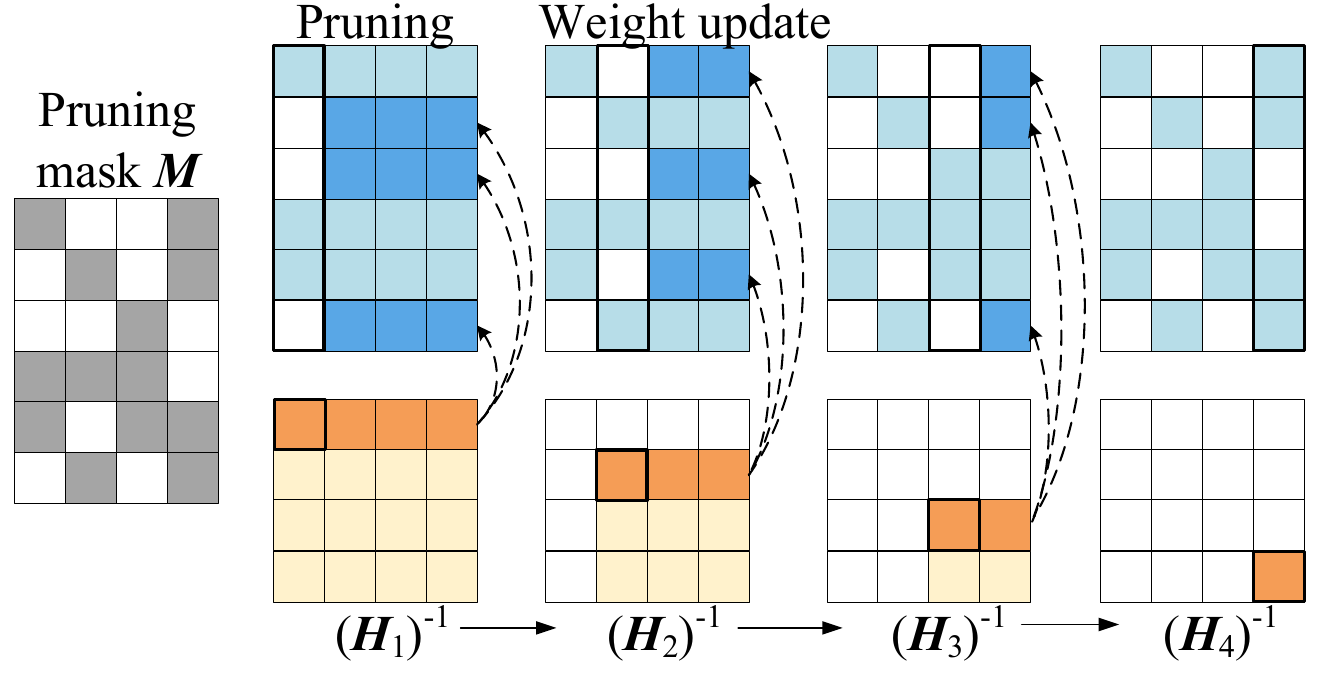}
	\caption{Pruning diagram}
	\label{fig:7}
\end{figure}

\noindent
To adapt to resource-constrained environments, reduce overfitting risks, and enhance model generalization, we employed the method proposed by Sun et al.\textsuperscript{\cite{53}}, named Sparesgpt, to optimize the structure and performance of the model. As illustrated in Figure \ref{fig:7}, the pruning process involves mask selection and weight reconstruction. The pruning mask \(\bm{M}\) is a binary matrix used to indicate which weights should be retained or pruned, and \(\bm{H}\) is the matrix used for optimizing weights. During model pruning, some weights are zeroed out or removed. To preserve model effectiveness, weights are reallocated during pruning, with updated weights assigned to the pruned (deep) portion and unpruned weights remaining unaltered (light).
\noindent
The Hessian matrix is employed by Sparsegpt to guide pruning decisions. The calculation is as shown in equation \eqref{17}. The error introduced after removing the weight at index \( m \) is calculated as shown in equation \eqref{18}, and the weight update calculation is shown in equation \eqref{19}.
\vskip -3mm
\begin{equation}
	\boldsymbol{H}_{\boldsymbol{M}_i}=\boldsymbol{X}_{\boldsymbol{M}_i} \boldsymbol{X}^{\mathrm{T}}{ }_{\boldsymbol{M}_i}\label{17}
\end{equation}
\begin{equation}
	\varepsilon_m=\frac{\omega_m^2}{\left|\boldsymbol{H}^{-1}\right|_{m, m}}\label{18}
\end{equation}
\begin{equation}
	\delta_m=-\frac{\omega_m}{\left|\boldsymbol{H}^{-1}\right|_{m, m}} \cdot \boldsymbol{H}_{:, m}^{-1}\label{19}
\end{equation}
Where $\boldsymbol{H}_{\boldsymbol{M}_i}$ represents the hessian matrix associated with the subset of model parameters that correspond to the pruning mask \( \bm{M}_i \), \( \boldsymbol{X}_{\boldsymbol{M}_i} \) represents a subset of input features corresponding to the pruning mask \( \bm{M}_i \), \( w_m \) is the weight, $\varepsilon_m$ represents the error introduced into the model as a result of removing the weight at index $m$, $\delta_m$  is the update to be applied to the weight at index $m$ during the pruning process.
\subsubsection{Loss function}
\noindent
The loss function \( L_{\text{ACOS}} \) for the algorithm, as shown in equation \eqref{20}, is comprised of three modules of loss, which are the aspect-opinion expression loss (\( L_{\text{AOE}} \)), sentiment classification loss (\( L_{\text{SP}} \)), and aspect category loss (\( L_{\text{AC}} \)). \( L_{\text{AOE}} \) is the CRF loss, as shown in equation \eqref{21}.
\begin{equation}
	L_{\operatorname{ACOS}} = L_{\operatorname{AOE}} + L_{\operatorname{SP}} + L_{\operatorname{AC}}\label{20}
\end{equation}
\begin{equation}
	L_{\operatorname{AOE}}=-\sum_{i=1}^N \log P\left(y^{(i)} \mid x^{(i)} ; \lambda\right)\label{21}
\end{equation}
Where \( N \) is the number of samples, \( x^{(i)} \) is the input sequence of the \( i \)-th sample, \( y^{(i)} \) is the corresponding true label sequence, \( P \) is the conditional probability of label sequence \( y \) given input sequence \( x \) and model parameters \( \lambda \). Both \( L_{\text{SP}} \) and \( L_{\text{AOE}} \) are cross-entropy losses. \( L_{\text{SP}} \) is represented as shown in equation \eqref{22}, while \( L_{\text{AOE}} \) is represented as shown in equation \eqref{23}.
\begin{equation}
	L_{\operatorname{SP}}=-\frac{1}{N} \sum_{i=1}^N \sum_{j=1}^{K_1} \hat{y}_{ij} \log \left(y_i\right)+\omega_1\left\|\theta_1\right\|^2\label{22}
\end{equation}
\begin{equation}
	L_{\operatorname{AC}}=-\frac{1}{N} \sum_{i=1}^N \sum_{j=1}^{K_2} \hat{y}_{ij} \log \left(y_i\right)+\omega_2\left\|\theta_2\right\|^2\label{23}
\end{equation}
Where \( N \) represents the number of samples, \( K_1 \) represents the aspect category to which the aspect term is assigned, \( K_2 \) represents the sentiment category to which the aspect term belongs, \(\hat{y}_{ij}\) represents the predicted probability that the \(i\)-th sample belongs to category \(j\), \( y_i \) is the corresponding true label, \( \omega \)  is the weight coefficient for the regularization term, and  \( \| \theta \|^{2} \) is the regularization term.

\section{Experimental results and analysis}
\subsection{Datasets description}
\noindent
Tourist review data from selected tourist attractions in Xinjiang is collected from Ctrip and Mafengwo websites. Emotion quadruples are extracted using ChatGPT, and to ensure accuracy, they are manually corrected. Experiments are conducted on two publicly available restaurant review datasets, Rest15 and Rest16, to verify the model's generalization and robustness. The statistical distribution of samples in the datasets is shown in Table \ref{tab:1}, which the columns labeled ``Positive'', ``Negative'', and ``Neutral'' represent the respective counts of positive, negative, and neutral quadruples.
\begin{figure*}[ht]
	\centering
	\subfigure[Travel]{
		\begin{minipage}[t]{0.3\linewidth}
			\centering
			\includegraphics[width=5cm,height=5cm]{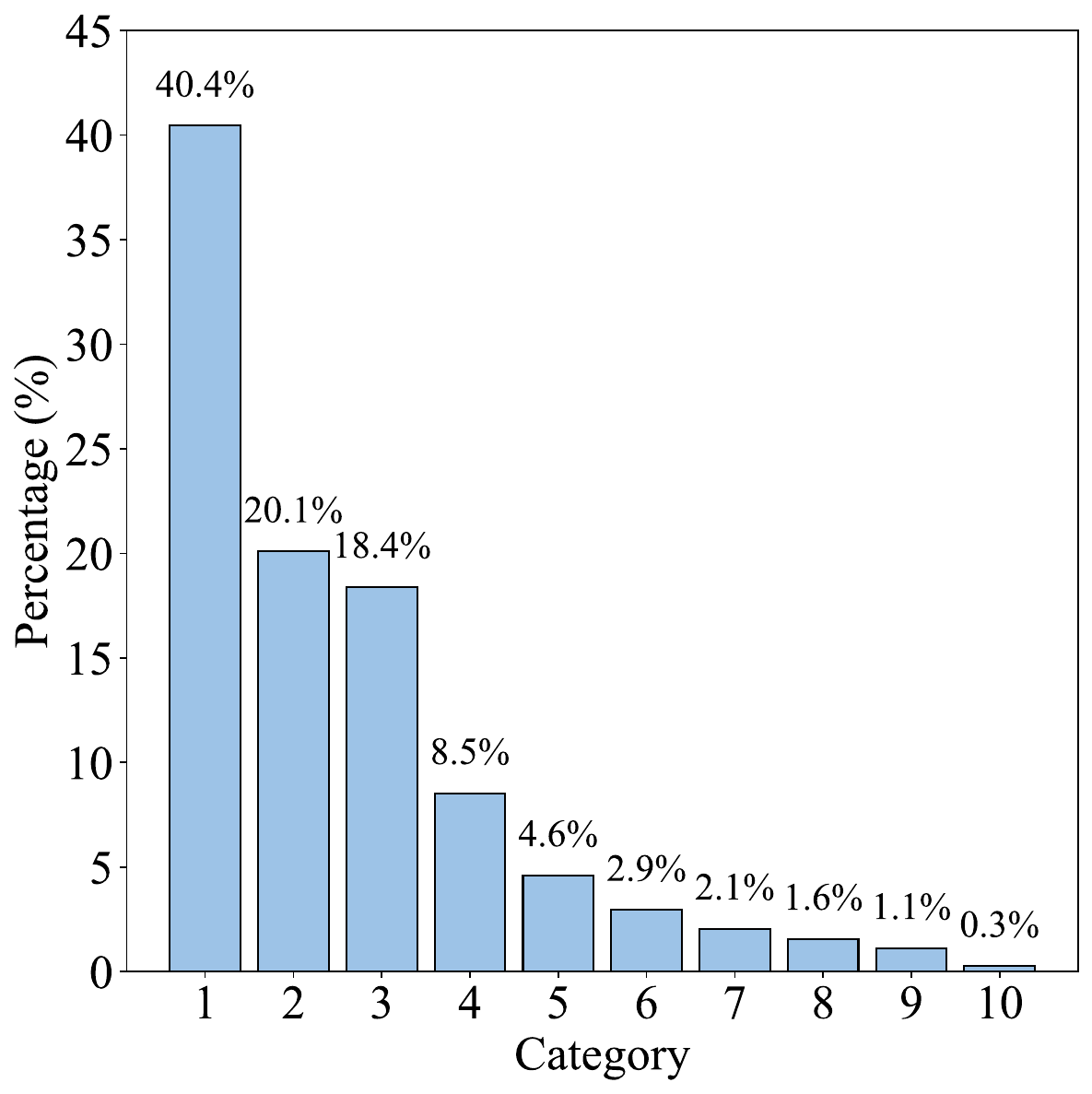}
		\end{minipage}
	}
	\hspace{0.1cm} 
	\subfigure[Rest15]{
		\begin{minipage}[t]{0.3\linewidth}
			\centering
			\includegraphics[width=5cm,height=5cm]{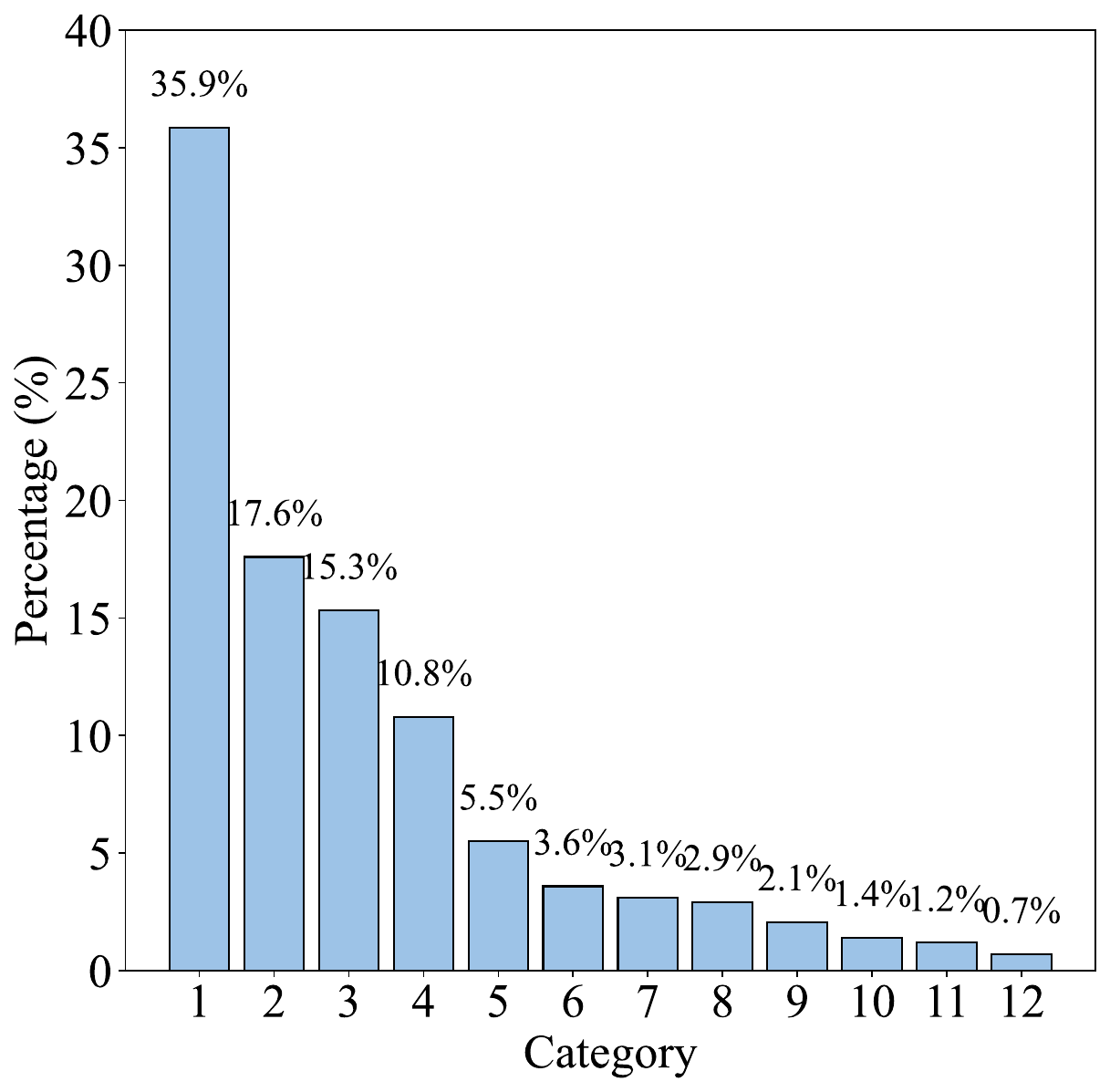}
		\end{minipage}
	}
	\hspace{0.1cm}
	\subfigure[Rest16]{
		\begin{minipage}[t]{0.3\linewidth}
			\centering
			\includegraphics[width=5cm,height=5cm]{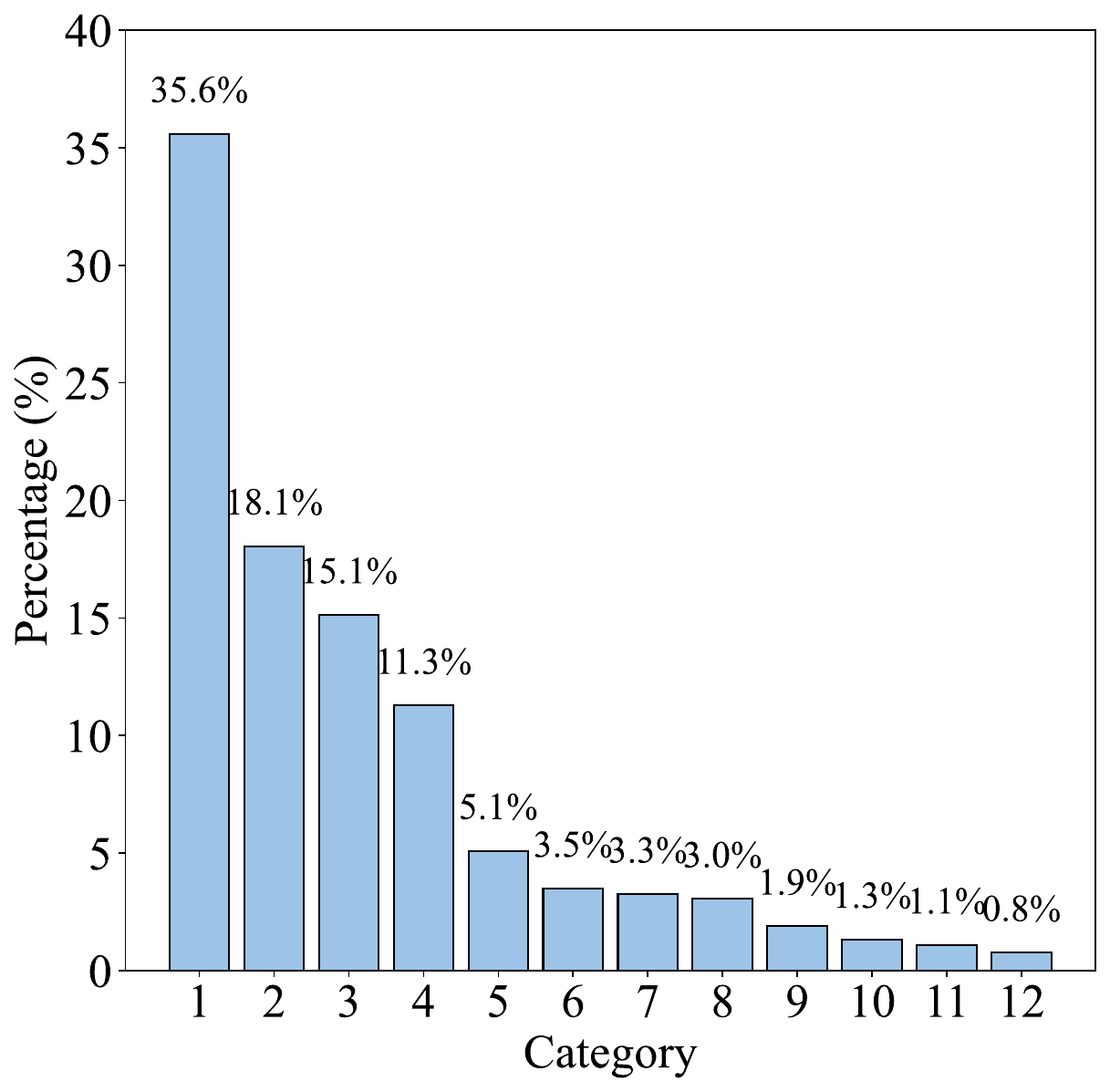}
		\end{minipage}
	}
	\vspace{-3mm} 
	\setcounter{subfigure}{0}
	\caption{Data count statistics for different categories}
	\label{fig:8}
\end{figure*}
\begin{table}[htbp]
	
	\centering
	\setlength{\tabcolsep}{2.2pt} % Set the horizontal padding
	\vspace{-2mm}
	\caption{Statistics of the datasets for ABSA tasks}
	\vspace{3mm}
	\begin{tabular}{ccccccc}
		\toprule
		Dataset & Split & Sentence & Quadruple & Positive & Negative & Neutral \\
		\midrule
		& Train & 1699 & 2024 & 842 & 955 & 227 \\
		Travel & Test & 394 & 475 & 107 & 283 & 85 \\
		& All & 2093 & 2499 & 949 & 1238 & 312 \\
		& Train & 1043 & 2951 & 2048 & 797 & 106 \\
		Rest15 & Test & 537 & 343 & 243 & 81 & 19 \\
		& All & 1580 & 3294 & 2291 & 878 & 125 \\
		& Train & 1893 & 1701 & 1257 & 48 & 396 \\
		Rest16 & Test & 230 & 795 & 453 & 305 & 37 \\
		& All & 2123 & 2496 & 1710 & 353 & 433 \\
		\bottomrule
	\end{tabular}
	\label{tab:1}
\end{table}
\par
Figure \ref{fig:8} illustrates the data count statistics for each category in the dataset. The horizontal axis represents different categories (replaced with data instead of category names), while the vertical axis represents the data count. In the tourism dataset, the comparison of different category quadruples is illustrated in Figure \ref{fig:8}(a), where the categories represented by 1 to 10 correspond to \{``service'', ``management'', ``environment'', ``price'', ``transportation'', ``attraction'', ``scenery'', ``cuisine'', ``facility'', ``project''\}. In Rest15 and Rest16, the comparison of different category quadruples is depicted in Figure \ref{fig:8}(b) and Figure \ref{fig:8}(c), where the categories represented by 1 to 10 correspond to \{``food quality'', ``restaurant general'', ``service general'', ``ambience general'', ``restaurant miscellaneous'', ``food style options'', ``restaurant prices'', ``drinks quality'', ``food prices'', ``location general'', ``drinks style options'', ``drinks prices''\}.
\subsection{Evaluation metrics}
\noindent
To evaluate the actual effectiveness of the model, the $F1$ score is adopted as the evaluation metric, where $P$ represents precision and $R$ represents recall. The calculation method is as follows.
\begin{equation}
	P = \frac{\text{n\_tp}}{\text{ n\_pred}}\label{24}
\end{equation}
\begin{equation}
	R= \frac{\text{n\_tp}}{\text{ n\_gold}}\label{25}
\end{equation}
\begin{equation}
	F 1=\frac{2 \times P \times R}{P+R}\label{26}
\end{equation}
\noindent
n\_tp represents the number of emotion quadruples correctly predicted by the model,  where a prediction is deemed accurate only if all components of the quadruple align precisely with the true labels. n\_pred represents the total number of emotion quadruples predicted by the model, and n\_gold represents the total number of emotion quadruples in the true labels.
\subsection{Experiments setting }
\noindent
The experiment is conducted using the Windows operating system, with Python version 3.9. The experimental environment utilizes Transformers version 4.34.1 released by Hugging Face, and the parameter settings are as shown in Table \ref{tab:2}. In the baseline experiments, models based on Llama2-7B and ACOS\_LLM use an A40 (48GB) GPU, while other models use a V100 (32GB).
\begin{table}[h]
	\centering
	\vspace{-2mm}
	\caption{Parameter settings}
	\vspace{3mm}
	\resizebox{\linewidth}{!}{%
		\begin{tabular}{ccc}
			\toprule
			Module & Parameters & Value \\
			\midrule
			\multirow{5}{*}{\centering \shortstack{The stage of\\ auxiliary knowledge\\ generation}} & Init\_r & 6 \\
			& Target\_r & 4 \\
			& Learning rate & 2e-4 \\
			& Batch size & 2 \\
			& Epoch & 30 \\
			\midrule
			\multirow{7}{*}{\centering \shortstack{The stage\\ of ACOSQE}} & SRD & 3 \\
			& Learning\_rate & 1e-4 \\
			& Optimizer & Adam \\
			& Batch size & 32 \\
			& embedding dimension & 768 \\
			& Loss function & Cross Entropy \\
			& Epoch & 100 \\
			\bottomrule
			\vspace{-6mm}
		\end{tabular}%
		\vspace{-2mm}
	}
	\label{tab:2}
\end{table}
\subsection{Main results}
\noindent
The model structure is comprised of two key stages: auxiliary knowledge generation and ACOSQE. Each part of the model is trained separately, with the output of the first-stage model being used as the input for the second-stage model. The extraction results of the second part quadruples are used to evaluate the model performance.
\begin{figure*}[t]
	\centering
	\subfigure[Travel]{
		\begin{minipage}[t]{0.3\linewidth}
			\centering
			\includegraphics[width=5.2cm,height=4.2cm]{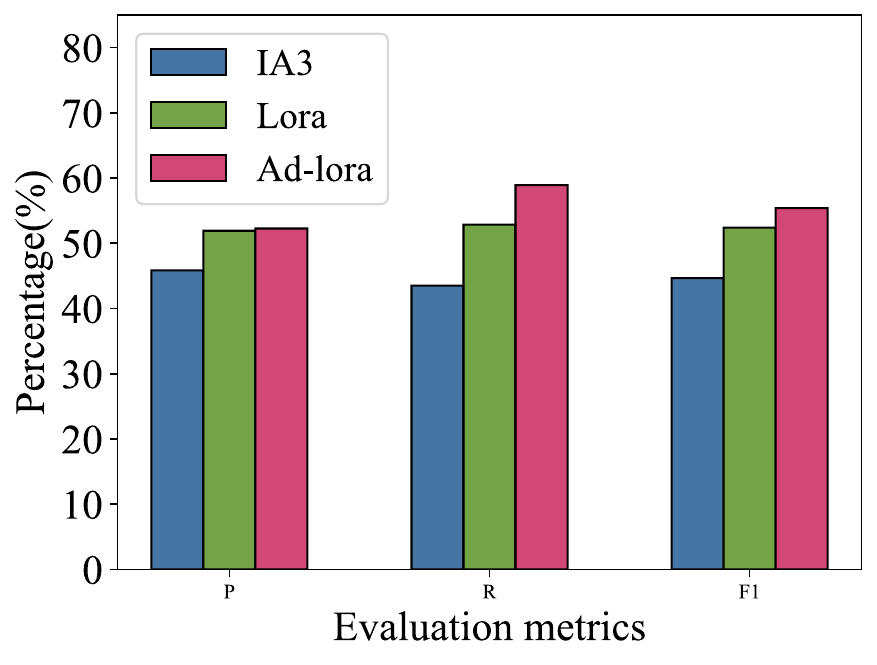}
		\end{minipage}
	}
	\hspace{0.1cm}
	\subfigure[Rest15]{
		\begin{minipage}[t]{0.3\linewidth}
			\centering
			\includegraphics[width=5.2cm,height=4.2cm]{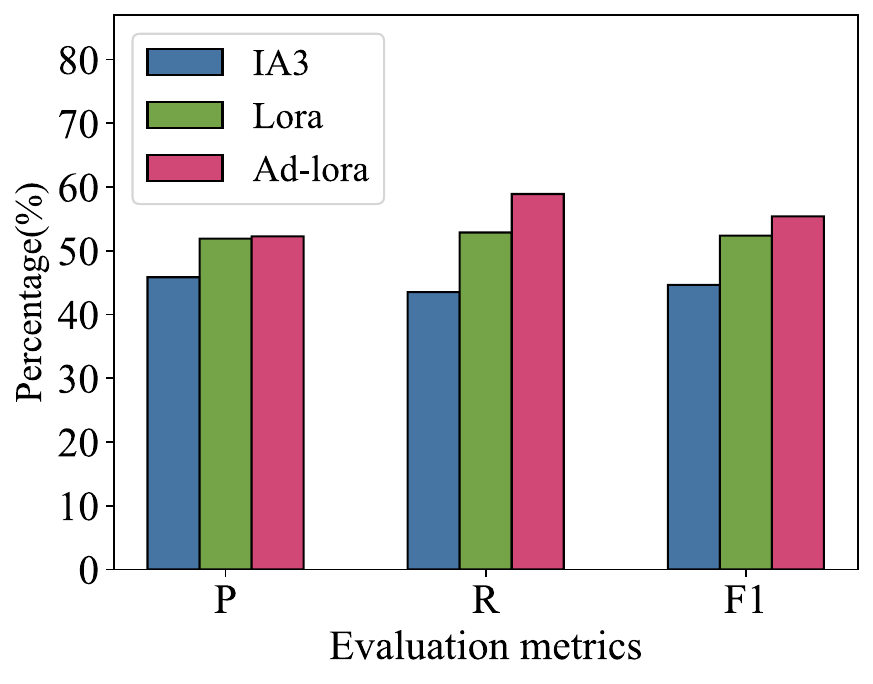}
		\end{minipage}
	}
	\hspace{0.1cm} 
	\subfigure[Rest16]{
		\begin{minipage}[t]{0.3\linewidth}
			\centering
			\includegraphics[width=5.2cm,height=4.2cm]{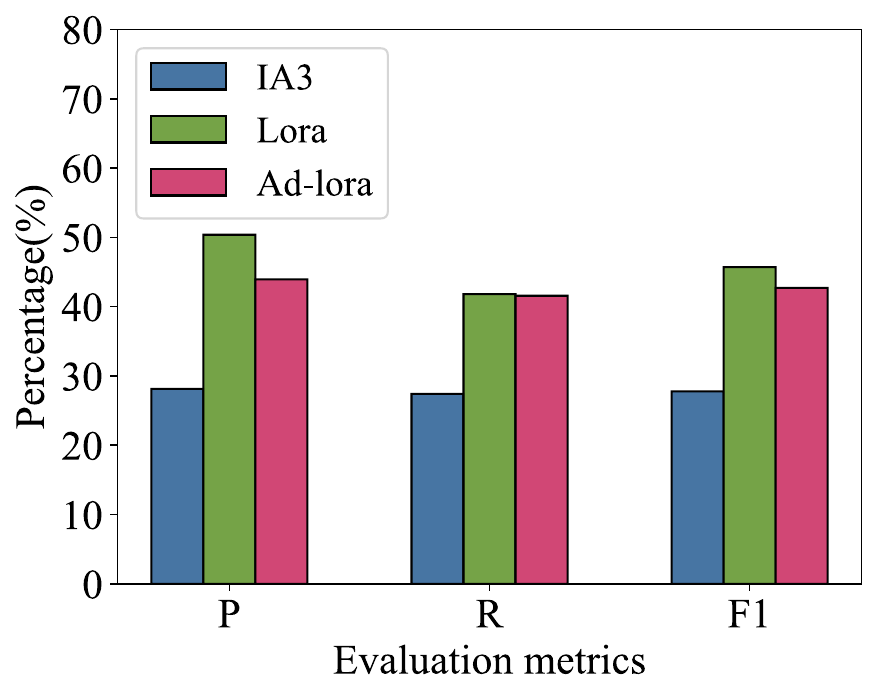}
		\end{minipage}
	}
	\vspace{-2.5mm}
	\setcounter{subfigure}{0}
	\caption{The influence of different fine-tuning methods on the model effect}
	\label{fig:9}
\end{figure*}
\begin{figure*}[ht]
	\centering
	\subfigure[Travel]{
		\begin{minipage}[t]{0.318\linewidth}
			\centering
			\includegraphics[width=5.2cm,height=4.2cm]{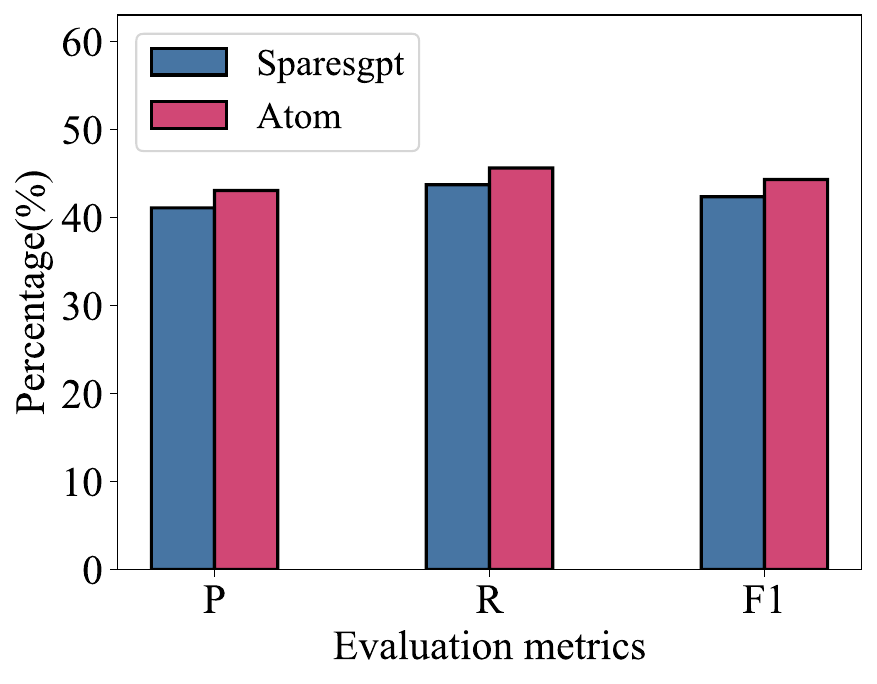}
			\vspace{-10mm}
		\end{minipage}
	}
	\hspace{-0.1cm} 
	\subfigure[Rest15]{
		\begin{minipage}[t]{0.318\linewidth}
			\includegraphics[width=5.2cm,height=4.2cm]{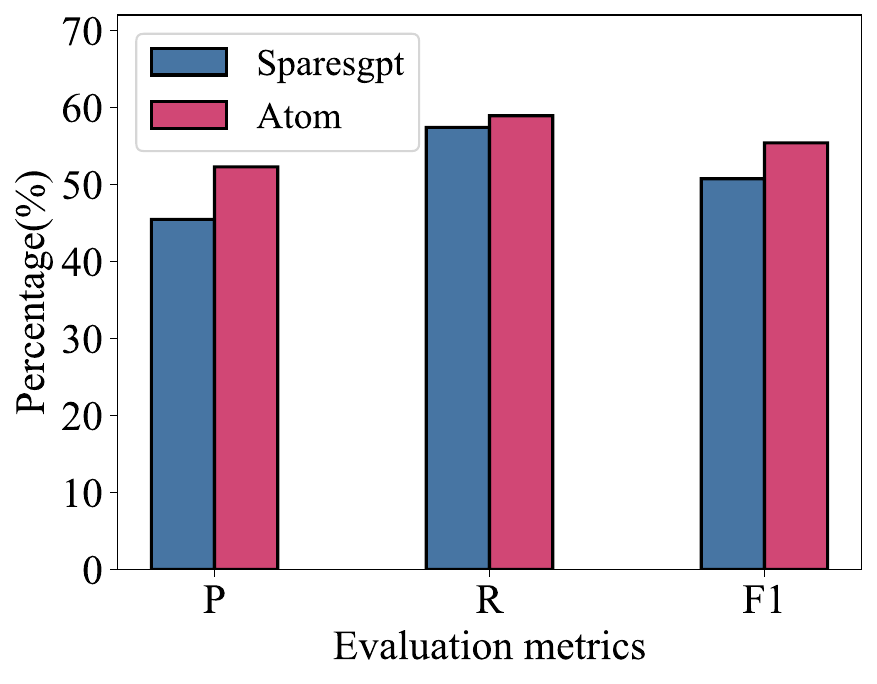}
			\vspace{3mm} 
		\end{minipage}
	}
	\hspace{-0.1cm}
	\subfigure[Rest16]{
		\begin{minipage}[t]{0.318\linewidth}
			\centering
			\includegraphics[width=5.2cm,height=4.2cm]{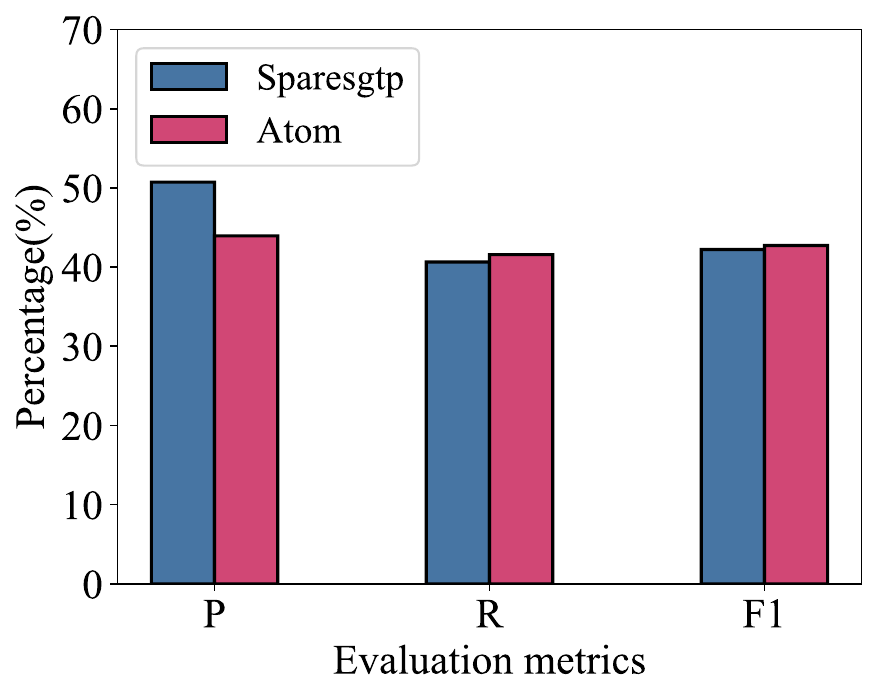}
			\vspace{3mm}
		\end{minipage}
	}
	\vspace{-3mm}
	\caption{Experimental results of model pruning}
	\label{fig:10}
\end{figure*}
\subsubsection{The stage of auxiliary knowledge generation}
\noindent
To effectively capture the semantic elements of sentiment and generate high-quality auxiliary knowledge, firstly, we conduct supervised fine-tuning on the model using labeled sentiment quadruples. The inputs are formatted as follows:
\par
Instruction: \{extract aspects and sentiment terms from the following text and classify sentiment polarity and evaluation category\}, input:\{text\}, output: \{sentiment elements\}<$s$>.
\par
IA3\textsuperscript{\cite{62}}, Lora, and Adalora are used to fine-tune Atom-7B, with their trainable parameter percentages being 0.008\%, 0.47\%, and 0.04\% respectively. The experimental results are shown in Figure \ref{fig:9}. Low-rank matrices \(\bm{A}\) and \(\bm{B}\) are integrated with the original weight matrices as side paths in Lora, where their product is used as trainable parameters to simulate parameter variations. AdaLora, a variant of Lora, employs SVD for low-rank decomposition instead of \(\bm{AB}\), resulting in higher parameter efficiency compared to Lora. In IA3, each activation layer is assigned a coefficient vector, which is multiplied with the activation units during both training and inference to reduce the number of parameters needing adjustment.
\par
From Figure \ref{fig:9}(a), it can be seen that efficient fine-tuning using Adalora significantly improves the model's ability. Compared to fine-tuning with Lora, F1 improves by 6.87\%, and compared to IA3 fine-tuning, F1 improves by 6.41\%. From Figure \ref{fig:9}(b), it can be seen that Lora performs the best in Rest15, with an F1 improvement of 2.99\% compared to other best-performing methods.  From Figure \ref{fig:9}(c), it can be seen that Adalora performs the best in Rest16, with an F1 improvement of 10.74\% compared to other best-performing methods.
\par
\begin{figure}[htb]
	\centering
	%\vspace{-3.4mm}/
	\includegraphics[width=0.9\linewidth]{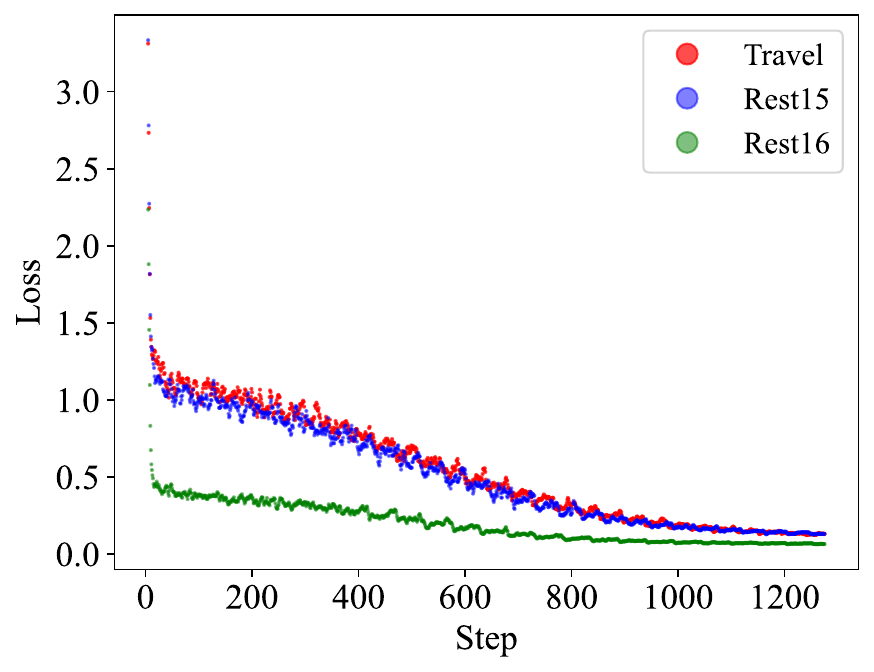}
	\caption{Change in loss value}
	\label{fig:11}
	\vspace{-3.4mm}
\end{figure}

Model pruning helps adapt the model to resource-constrained scenarios, improving efficiency and deployability. To analyze the impact of pruning on model performance, Sparsegpt is used to prune the model to a sparsity of 50\%. The results of the model before and after pruning are shown in Figure \ref{fig:10}. It can be observed that pruning the model to 50\% sparsity results in only a slight decrease in F1 score on the tourism dataset, with a decrease of only 1.95\%. On the Rest16, F1 decreases by 4.66\%, while on Rest15, the model pruning performs the best, with an increase in precision by 6.79\% and a decrease in F1 by only 0.51\%. Thus, it can be concluded that utilizing Sparsegpt for model pruning can achieve model compression while maintaining model performance. The loss changes of the models efficiently fine-tuned using Adalora are shown in Figure \ref{fig:11}. It can be observed that in the tourism dataset, Rest15, and Rest16, the loss values tend to stabilize when the iteration steps reach 10,000-12,000, with the model converging fastest on the Rest16.
\subsubsection{The stage of ACOSQE}
\begin{figure*}[htbp]
	\centering
	\subfigure[The influence of SRD on the model effect]{
		\begin{minipage}[t]{0.4\linewidth}
			\centering
			\includegraphics[width=6.2cm,height=5cm]{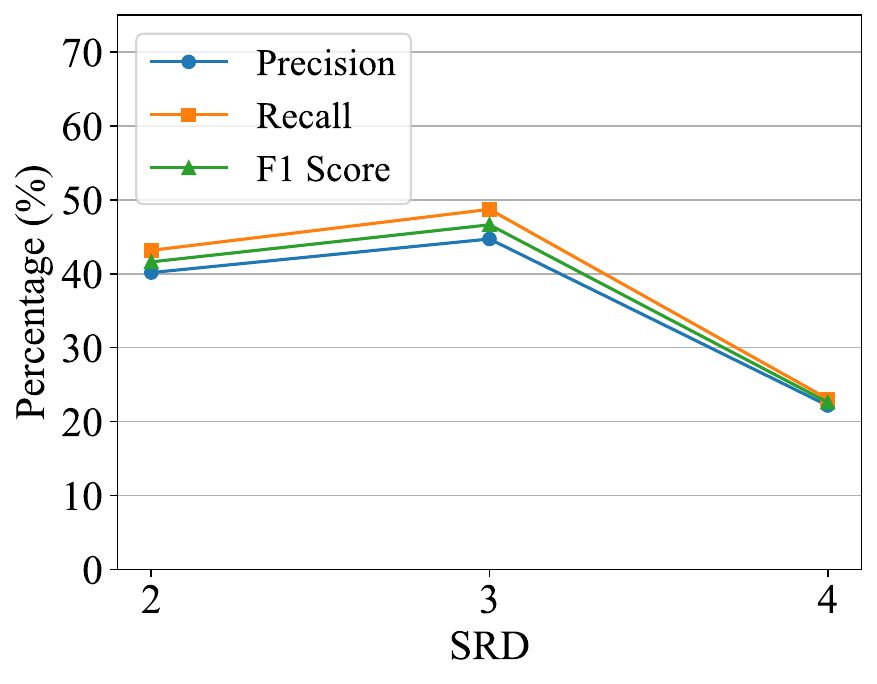}
		\end{minipage}
	}
	\hspace{1.1cm} 
	\subfigure[The influence of position information on the model effect]{
		\begin{minipage}[t]{0.4\linewidth}
			\centering
			\includegraphics[width=6.2cm,height=5cm]{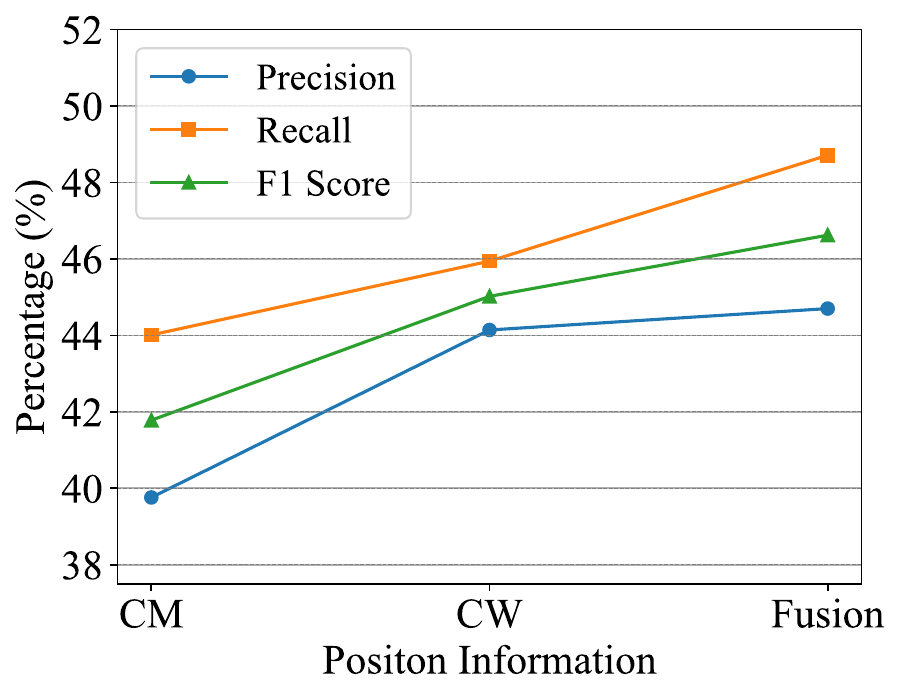}
		\end{minipage}
	}
	\vspace{-1mm} 
	\setcounter{subfigure}{0}
	\caption{The influence of SRD and position information on the model effect}
	\label{fig:12}
\end{figure*}

\noindent
The second stage of the proposed model involves joint training for Aspect-Opinion Pairs Extraction and aspect category sentiment classification. Considering that multiple aspect terms may exist in a sentence, in order to accurately capture the sentiment corresponding to each aspect term, different weights are assigned to the surrounding information of aspect terms based on their positional information. SRD represents the distance between words in the text and aspect terms, which allows controlling the model's attention to information at different distances when used as a threshold. To study the impact of SRD , experiments are conducted where SRD was set to 3, 4, and 5 respectively, while keeping other parameters unchanged. The experimental results are shown in Figure \ref{fig:12}. To enhance the model's ability to handle positional and contextual information, three information fusion methods are employed: CM, CW, and Fusion. CW and CM utilize mask features and weight features respectively, while Fusion combines the two features using concatenation. The experimental results are depicted in Figure \ref{fig:12}.
\par
Figure \ref{fig:12}(a) demonstrates that the model achieves optimal performance when SRD is set to 3. This observation can be attributed to the fact that a smaller SRD causes the model to concentrate predominantly on the contextual information proximal to the entity, thereby neglecting the broader textual context. An excessively large SRD might cause the model to miss crucial sentiment-related details associated with the entity, adversely affecting the sentiment classification efficacy. Figure \ref{fig:12}(b) illustrates that the best results are obtained using the Fusion method. It is due to the CM feature, which zeroes out information beyond the SRD threshold, resulting in a loss of some relevant data. Although the CW method preserves a greater amount of aspect-related information by applying weights, it concurrently introduces noise, which can detract from model accuracy.
\subsubsection{Baselines}
\noindent
To evaluate the performance of the model, the following models are used as benchmarks for comparison:
\begin{itemize}
	\item \textbf{LCF\_ATEPC}\textsuperscript{\cite{63}} is an aspect sentiment extraction model based on joint methods, which utilizes relative positional distance to assist sentiment polarity prediction.
	
	\item \textbf{Mt5-small} and \textbf{Mt5-base}\textsuperscript{\cite{64}} transform sentiment extraction into paraphrase generation questions, achieving end-to-end resolution. 
	
	\item \textbf{Mt5+GRU} adds a GRU layer to Mt5-base\textsuperscript{\cite{65}} to help the model capture long-term dependencies in sequences. Model parameters are updated using both GRU loss and Mt5-base loss. In order to balance their influences, the weight of the GRU loss is set to 0.05.
	
	\item \textbf{TASO}\textsuperscript{\cite{66}} simplifies the joint detection problem into a binary text classification task based on the presence of aspect-sentiment pairs and a sequence labeling task.
	
	\item \textbf{GTS}\textsuperscript{\cite{67}} captures the interaction dependency information among the three elements in a sentence through a GCN with attention mechanisms.
	
	\item \textbf{Extract-Classify-ACOS}\textsuperscript{\cite{68}} first extracts aspects and opinions, and predicts the implicit aspects and opinions, then performs ACOS classification.
		
	\item \textbf{Llama-7B}\textsuperscript{\cite{69}} utilizes Adalora for fine-tuning and then directly extracts emotion quadruples from the text through generation. The parameter settings for model training are as follows: initial rank is 6, target rank is 4, learning rate is 0.0002, batch size is 2, and the training lasts for a total of 30 epochs.
		
	\item \textbf{ACOS\_LLM} utilizes auxiliary knowledge and location information to assist in quadruple emotional extraction, with Adalora and Sparsegpt being used to optimize the generation of auxiliary knowledge by Atom-7b.
\end{itemize}
\par
The experiments are conducted on a self-constructed tourism dataset and public dataset. As shown in Table \ref{tab:3}, to validate the performance of the proposed model on other ABSA tasks, tests are conducted on Category-Sentiment Pair Extraction (CS), Target-Aspect-Sentiment Triplet Extraction (TASD), and Aspect-Sentiment-Term Extraction (ASTE) tasks, respectively. The last two columns display the number of parameters and the inference speed (measured by the number of sentences processed per second). For ACOS\_LLM, only the parameter count for the quadruple extraction module is listed, while the auxiliary information generation part has 7 billion parameters.
\begin{table*}[hbt]
	\centering
	\caption{Performance comparison (Results with * are copied from the raw papers, bold numbers indicate the best-performing model for F1 score)}
	\vspace{7pt}
	\resizebox{\textwidth}{!}{
		\vspace{3mm}
		\begin{tabular}{cccccccccccccccc}
			\toprule
			\multirow{2}{*}{Dataset} & \multirow{2}{*}{Model} & \multicolumn{3}{c}{ACOS} & \multicolumn{3}{c}{CS} & \multicolumn{3}{c}{TASD} & \multicolumn{3}{c}{ASTE} & \multirow{2}{*}{Para} & \multirow{2}{*}{Speed}\\
			\cmidrule(lr){3-5} \cmidrule(lr){6-8} \cmidrule(lr){9-11} \cmidrule(lr){12-14}
			&  & P & R & F1 & P & R & F1 & P & R & F1 & P & R & F1 &  & \\
			\midrule
			\multirow{5}{*}{Travel} 
			& LCF\_ATEPC & 37.53 & 30.55 & 33.68 & 57.34 & 42.52 & 48.83 & 45.26 & 36.75 & 40.56 & 46.41 & 37.39 & 41.42 & 180M & 51.3 \\
			& Mt5-small & 29.49 & 29.75 & 29.62 & 56.69 & 57.17 & 56.93 & 41.42 & 41.76 & 41.59 & 35.98 & 36.28 & 36.13 & 300M & 103.1 \\
			& Mt5-base & 28.13 & 29.32 & 28.71 & 52.63 & 54.85 & 53.71 & 40.08 & 41.77 & 40.90 & 35.42 & 36.91 & 36.15 & 582M & 65.9 \\
			& Mt5+GRU & 31.18 & 31.64 & 31.41 & 54.26 & 55.09 & 54.65 & 41.99 & 42.61 & 42.30 & 38.04 & 38.60 & 38.32 & 583M & 63.1 \\
			& Llama-7B & 37.84 & 40.51 & 39.13 & 61.75 & 66.09 & \textbf{63.85} & 50.99 & 54.58 & 52.72 & 47.21 & 50.53 & 48.81 & 7B & 3.8 \\
			& ACOS\_LLM & 44.70 & 48.71 & \textbf{46.62} & 61.22 & 61.75 & 61.48 & 50.99 & 54.91 & \textbf{52.88} & 53.67 & 56.19 & \textbf{54.90} & 181M & 12.4 \\
			\midrule
			\multirow{7}{*}{Rest15} 
			& Mt5-small & 36.29 & 36.47 & 36.38 & 64.70 & 65.03 & 64.86 & 53.44 & 53.71 & 53.57 & 44.30 & 44.52 & 44.41 & 300M & 101.4 \\
			& Mt5-base & 40.26 & 38.99 & 39.61 & 64.67 & 62.64 & 63.64 & 55.58 & 53.83 & 54.69 & 50.12 & 48.55 & 49.32 & 582M & 64.1 \\
			& Mt5+GRU & 41.77 & 39.62 & 40.67 & 68.70 & 65.15 & 66.88 & 57.95 & 54.96 & 56.42 & 50.79 & 48.17 & \textbf{49.45} & 583M & 64 \\
			& Extract-Classify-ACOS* & 35.64 & 37.25 & 36.42 & - & - & - & - & - & - & - & - & - & - & - \\
			& TASO* & 44.24 & 28.66 & 34.74 & - & - & 70.42 & - & - & 58.09 & - & - & - & - & - \\
			& GTS* & 43.13 & 42.18 & 42.67 & - & - & 65.45 & - & - & 57.76 & - & - & - & - & - \\
			& Llama-7B & 38.01 & 42.85 & 40.29 & 70.89 & 79.92 & 75.13 & 57.19 & 64.47 & 60.61 & 44.97 & 50.70 & 47.67 & 7B & 3.7 \\
			& ACOS\_LLM & 43.94 & 41.57 & \textbf{42.72} & 78.09 & 73.87 & \textbf{75.92} & 63.80 & 60.36 & \textbf{62.03} & 47.58 & 49.42 & 48.48 & 181M & 47.2 \\
			\midrule
			\multirow{7}{*}{Rest16} 
			& Mt5-small & 51.88 & 51.56 & 51.72 & 75.18 & 74.71 & 74.95 & 65.23 & 64.83 & 65.03 & 59.69 & 59.32 & 59.51 & 300M & 95.6 \\
			& Mt5-base & 49.63 & 51.68 & 50.64 & 73.31 & 76.34 & 74.80 & 64.42 & 67.08 & 65.72 & 58.05 & 60.45 & 59.22 & 582M & 61.2 \\
			& Mt5+GRU & 52.42 & 54.19 & 53.29 & 75.54 & 78.09 & 76.80 & 65.61 & 67.83 & 66.70 & 59.20 & 61.20 & 60.18 & 583M & 59.6 \\
			& Extract-Classify-ACOS* & 38.40 & 50.93 & 43.77 & - & - & - & - & - & - & - & - & - & - & - \\
			& TASO* & 48.65 & 39.68 & 48.65 & - & - & 76.33 & - & - & 65.44 & - & - & - & - & - \\
			& GTS* & 50.38 & 45.81 & 47.16 & - & - & 75.30 & - & - & 67.66 & - & - & - & - & - \\
			& Llama-7B & 52.07 & 56.79 & 54.33 & 77.00 & 83.98 & 80.34 & 65.65 & 71.60 & 68.49 & 54.84 & 59.81 & 57.22 & 7B & 3.5 \\
			& ACOS\_LLM & 52.27 & 58.91 & \textbf{55.39} & 77.21 & 87.00 & \textbf{81.81} & 65.14 & 73.41 & \textbf{69.03} & 56.83 & 64.04 & \textbf{60.22} & 181M & 44 \\
			\bottomrule
			\vspace{-8.5mm}
	\end{tabular}}
	\label{tab:3}
\end{table*}
\par
From Table \ref{tab:3}, ACOS\_LLM achieves the highest performance in the ACOS task, with an F1 score of 46.62\%, which improves by 7.49\% compared to the best-performing model among the comparison models. Additionally, the F1 scores for the TASD and ASTE tasks are the best compared to other models, with improvements of 0.16\% and 6.09\%, respectively. These results underscore the effectiveness of the proposed model across aspect term extraction, sentiment term extraction, and aspect-sentiment classification tasks. This is because aspect term and sentiment term extraction are assisted by auxiliary information, while sentiment classification is aided by distance information. However, in the CS task, the result decreases by 2.37\% compared to the best-performing model Llama-7B. This is due to the fact that aspect terms are mostly surrounded by sentiment terms, leading to less contribution to aspect category classification tasks, particularly in Chinese, where ambiguous expressions are common.
\par
LCF\_ATEPC, designed for unified information extraction and text classification tasks, ensures label accuracy during training, resulting in better performance than the Mt5 series algorithms. Compared to Mt5+GRU, it achieves a 2.27\% improvement in F1 score. Mt5\_base and Llama-7B transform the emotion extraction into a generation problem, providing a unified framework for aspect-level sentiment analysis, which are adaptable to various tasks and have low annotation costs. Llama-7B surpasses Mt5\_base by 10.42\% in F1 score due to its proficiency as a LLM, enabling better capture of linguistic features associated with sentiment information. However, the advantage of using Mt5 for sentiment extraction lies in its low resource requirements, which can be further enhanced by improving its structure. For example, incorporating GRU to capture long-term dependencies in sequences results in a 2.7\% improvement in F1 score compared to Mt5\_base, demonstrating the effectiveness of this approach.
\par
Additionally, to further validate the effectiveness of the proposed model, it is essential to investigate its generalization capabilities on other public datasets. The experimental results obtained after fine-tuning on the Rest15 and Rest16 datasets are shown in Table \ref{tab:3}, with hyperparameter settings consistent with Table \ref{tab:2}. The results indicate that in the ACOSQE task, compared to the best-performing models among the comparison models, our model achieves F1 score improvements of 0.05\% and 1.06\% on Rest15 and Rest16, respectively. Furthermore, Table \ref{tab:3} demonstrates that on Rest16, compared to other models, the proposed model achieves F1 values improved by 1.47\%, 0.54\%, and 0.04\% in the CS, TASD, and ASTE tasks, respectively. On Rest15, the proposed model also performs well, with F1 values improved by 0.79\% and 1.42\% in the CS and TASD tasks, compared to other models. However, in the ASTE task, the F1 value decreased by 0.97\% compared to the best-performing model, indicating that relevant hyperparameters need to be adjusted based on the characteristics and syntactic features of different datasets to optimize the model. Overall, the experimental results demonstrate the applicability of ACOS\_LLM in ABSA subtasks, and also showcase its generalization ability across various review datasets.
\subsubsection{Ablation experiment}
\noindent
To analyze the effectiveness of the proposed modules, ablation experiments are conducted in this section. The variant models designed for this purpose are listed below, and the results are shown in Table \ref{tab:4}.
\begin{itemize}
	\item \textbf{-ACOSQE Model}: The emotion quadruple extraction module is removed, and the task is achieved directly using Atom-7B as a text generation method for aspect representation.
	\item \textbf{-Knowledge}: The auxiliary knowledge generated by Atom-7B is removed, and the task is completed directly using the ACOSQE module.
	\item \textbf{-BERT}: BERT weights are frozen during parameter updates, and only downstream models are trained.
	\item \textbf{-LSTM}: In the Aspect-Opinion Pairs Extraction phase, the LSTM module is removed.
	\item \textbf{-Neural ODE}: In the Aspect-Opinion Pairs Extraction phase, the Neural ODE module is removed.
	\item \textbf{-GRU}: In the Aspect Category Sentiment Classification phase, the GRU module is removed.
\end{itemize}
\begin{table*}[htb]
	\centering
	\caption{The results of the ablation experiment (bold numbers indicate the best-performing model for F1 score)}
	\vspace{3mm}
	\resizebox{\textwidth}{!}{
	\begin{tabular}{ccccccccccccc}
		\toprule
		\multirow{2}{*}{Model} & \multicolumn{3}{c}{ACOS} & \multicolumn{3}{c}{CS} & \multicolumn{3}{c}{TASD} & \multicolumn{3}{c}{ASTE} \\
		\cmidrule(r){2-4} \cmidrule(lr){5-7} \cmidrule(lr){8-10} \cmidrule(l){11-13}
		& P & R & F1 & P & R & F1 & P & R & F1 & P & R & F1 \\
		\midrule
		-ACOS Model & 43.05 & 45.62 & 44.30 & 66.73 & 73.28 & \textbf{69.85} & 52.94 & 56.07 & \textbf{54.45} & 48.69 & 51.59 & 50.10 \\
		-Knowledge & 37.53 & 30.55 & 33.68 & 57.34 & 42.52 & 48.83 & 45.26 & 36.75 & 40.56 & 46.41 & 37.39 & 41.42 \\
		-BERT & 27.43 & 25.85 & 26.62 & 54.74 & 48.07 & 51.19 & 43.99 & 41.45 & 42.68 & 36.29 & 33.11 & 34.63 \\
		-BiLSTM & 37.06 & 41.02 & 38.94 & 54.71 & 57.05 & 55.85 & 44.83 & 49.14 & 46.89 & 48.27 & 50.85 & 49.53 \\
		-Neural ODE & 39.44 & 42.64 & 40.98 & 60.94 & 65.88 & 63.31 & 49.70 & 53.73 & 51.63 & 45.75 & 49.46 & 47.54 \\
		-BiGRU & - & - & - & - & - & - & - & - & - & 21.10 & 24.57 & 22.70 \\
		ACOS\_LLM & 44.70 & 48.71 & \textbf{46.62} & 61.22 & 61.75 & 61.48 & 50.99 & 54.91 & 52.88 & 53.67 & 56.19 & \textbf{54.90} \\
		\bottomrule
		\vspace{-8.5mm}
	\end{tabular}}
	\label{tab:4}
\end{table*}
\par
These variants are subjected to ablation experiments, and the results are presented in Table \ref{tab:4}. The results of the ablation experiments indicate that freezing BERT parameters and BiGRU have a significant impact on model performance. This is because freezing BERT parameters restricts the model from accessing information relevant to tourism quadruple, thereby limiting its ability to leverage contextual information during task execution. The results from Table \ref{tab:4} suggest that even after removing BiGRU, the model retains some capability in the ASTE task due to the assistance provided by auxiliary information in identifying aspect and sentiment words. However, due to the complexity of aspect category classification tasks, the original model utilizes BiGRU to capture sequence information to aid sentiment and category classification. Removing BiGRU results in a significant decline in classification capability, making it nearly impossible for the model to complete quadruplet extraction tasks.

Removing auxiliary knowledge resulted in a 12.94\% decrease in the model's F1 score in the ACOS task. This is because tourism ACOSQE task belongs to a low-resource scenario, where traditional training methods struggle to fully exploit semantic information due to limited data volume and also find it challenging to uncover implicit sentiment for aspect terms that are not explicitly mentioned in the text.  Therefore, utilizing large pre-trained models to generate auxiliary knowledge can provide rich semantic representations for the model, thereby alleviating the issue of insufficient knowledge representation in the tourism domain.
\par
\begin{figure}[htb]
	\vspace{4pt}
	\centering
	\vspace {-3.5mm}
	\includegraphics[width=\linewidth]{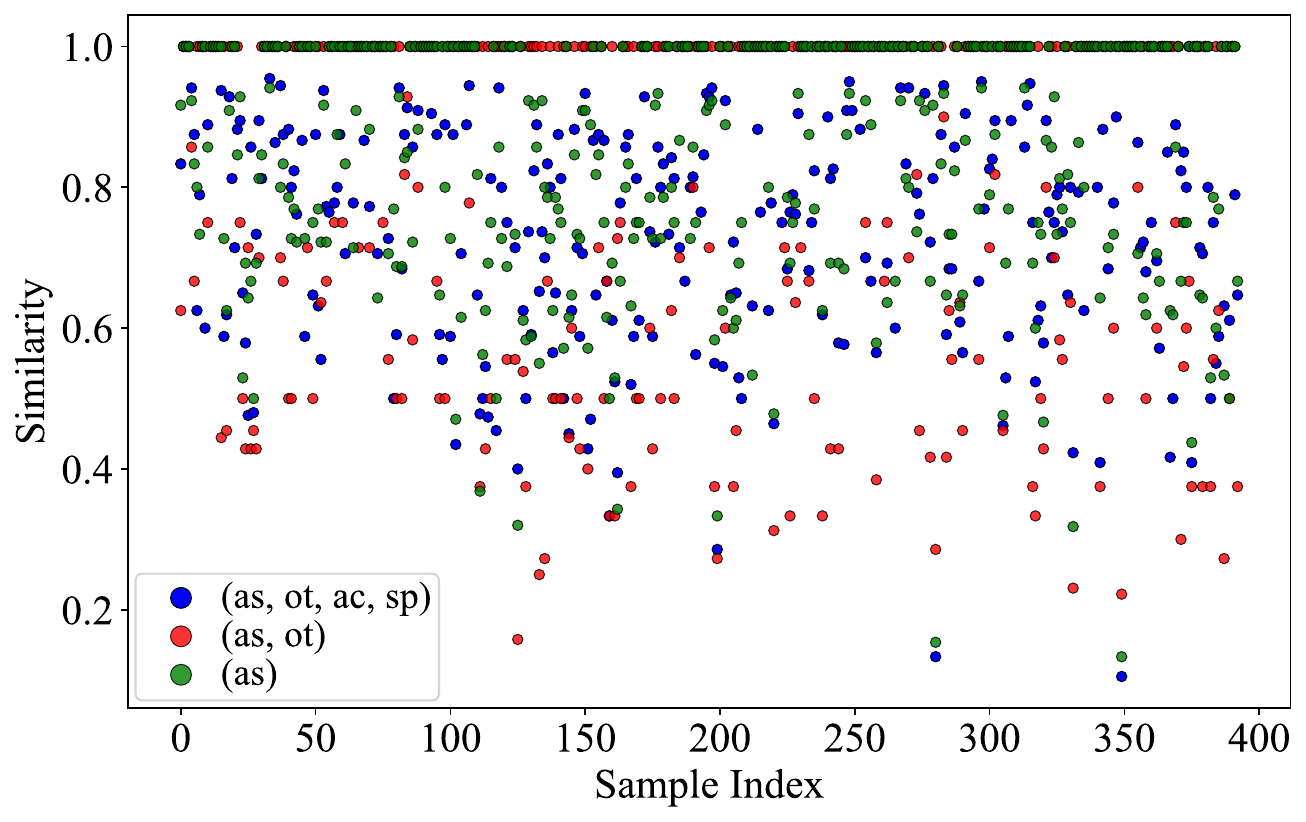}
	\vspace {-6.5mm}
	\caption{Analysis of the quality of auxiliary information}
	\vspace {-3mm}
	\label{fig:13}
\end{figure}

To analyze the impact of different levels of auxiliary information on the extraction of tourism sentiment quadruples, we compared the performance of sentiment quadruple extraction across models with varying auxiliary information: without auxiliary information (-data), with aspect information (+as), and with both aspect and opinion information (+(as, ot)). To validate the relevance of these auxiliary information types with true information, we employed a set-based similarity calculation method, the Jaccard similarity index, to assess the consistency between the generated auxiliary information and the correct information. The experimental results, as illustrated in the Figure \ref{fig:13}, show that although the auxiliary information generally correlates with the true information, it also introduces some noise.
\par
\vspace{7pt}
These auxiliary information types were compared against the baseline model without any auxiliary information to evaluate their actual effects, as shown in Table \ref{tab:5}. It can be concluded that the +as model increased the F1 score by 7.61\% compared to the -data model, as the inclusion of aspect information enhanced the model's ability to recognize specific aspects in the text, which in turn enabled accurate identification of corresponding sentiment polarities. Additionally, auxiliary information also increased the computational burden, resulting in a decrease of 7.2 in the number of texts processed per second. The +(as, ot) model improved the F1 score by 5.3\% over the +as model due to the added opinion information, which enhanced the model's ability to identify aspects and their sentiments with minimal impact on processing speed, effectively balancing efficiency and accuracy. Compared to the +(as, ot) configuration, the +(as, ot, ac, sp) configuration only showed a 0.05\% increase in F1 score, with a decrease of 1 in the number of texts processed per second, suggesting that the further added category and sentiment orientation information likely overlap highly with the already included aspect and opinion information, thus not significantly improving model performance.
\par
\begin{table}[h]
	\centering
	\vspace{-2.1mm}
	\caption{The impact of different auxiliary information on model performance (bold numbers indicate the best-performing model for F1 scores)}
	\vspace{1.5mm}
	\vspace{7pt}
	\resizebox{\linewidth}{!}{
	\begin{tabular}{ccccc}
		\toprule
		Model & P  & R  & F1  & Speed \\
		\midrule
		-data & 37.53 & 30.55 & 33.68 & 19.3 \\
		+as & 46.05 & 37.41 & 41.29 & 12.1 \\
		+(as, ot) & 43.61 & 50.00 & 46.59 & 13.1 \\
		+(as, ot, ac, sp) & 44.70 & 48.71 & \textbf{46.62} & 14.2 \\
		\bottomrule
	\end{tabular}
	\vspace{-1.5mm}
	\label{tab:5}
}
\end{table}
\par
Removing BiLSTM knowledge resulted in a 7.68\% decrease in the model's F1 score in the ACOS task. BiLSTM is employed to capture sequential information for identifying aspect, and the model determines sentiment polarity and aspect category based on these aspects in the second stage. Therefore, errors or omissions in entity recognition can lead to error accumulation and consequently affect quadruplet extraction.
\par
\begin{figure}[ht]
	\vspace{-2pt}
	\centering
	\includegraphics[width=1\linewidth]{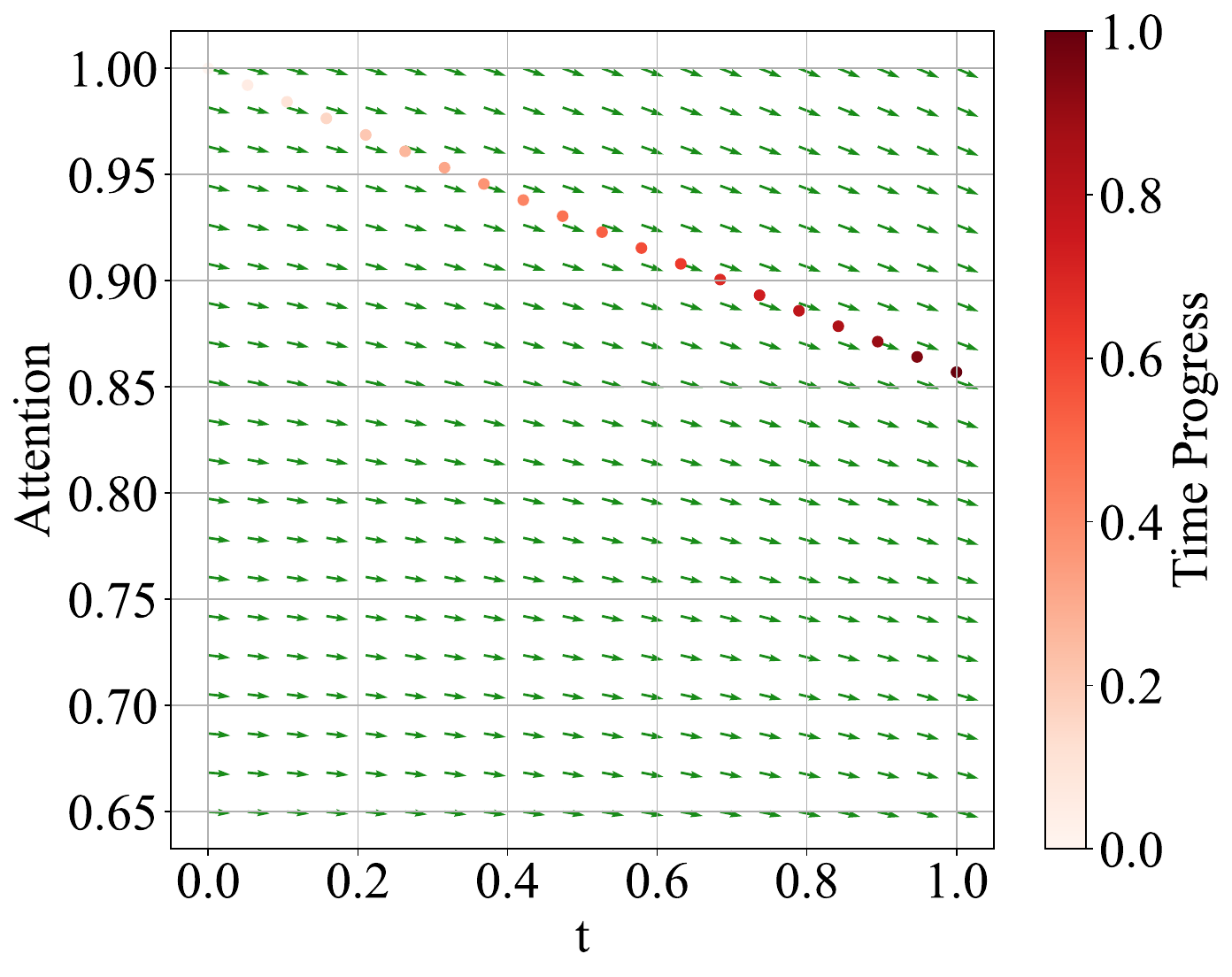}
	
	\caption{Continuity analysis}
	\vspace{-1.5mm}
	\label{fig:14}
\end{figure}

The ablation of Neural ODE modules affects model performance. ACOSQE establishes initial positional weights based on the relative positional information between aspect terms and the text, then employs Neural ODE to refine these positional attention weights. This method effectively captures the relationship between aspect terms and the overall sentence semantics, thus improving aspect-level sentiment classification. Using ``Tianshan is really beautiful, but the queue to get in is very chaotic'' as an example, ``Tianshan'' is assigned an initial weight of 1 as it is the aspect term. The evolution of attention weights implemented by a Neural ODE is depicted in Figure \ref{fig:14}. Arrows indicate the trend and speed of changes in attention weights at different time points and levels, with their direction and length representing these changes, and longer arrows suggesting faster changes. The dots represent specific numerical values in the attention weight evolution process, with colors transitioning from light to dark indicating the passage of time. It can be observed that the attention weight for ``Tianshan'' shows a decreasing trend over time.
\par
The ablation of ACOSQE modules also affects model performance to some extent. Although using a generation model to perform sentiment extraction tasks can improve the model's performance in handling text ambiguity and polysemy, the diversity in generated text often results in an expanded answer space, making it difficult to precisely match the real answers. The ACOSQE module, being a unified information extraction and text classification model, requires labels to correspond exactly with predicted results during training, which can compensate for the aforementioned shortcomings.
\subsubsection{Case study}
\begin{figure}[htb]
	\centering
	\subfigtopskip=2pt
	\subfigbottomskip=2pt
	\subfigcapskip=-5pt
	\subfigure[The weights before pruning]{
		\includegraphics[width=7.5cm,height=5.7cm]{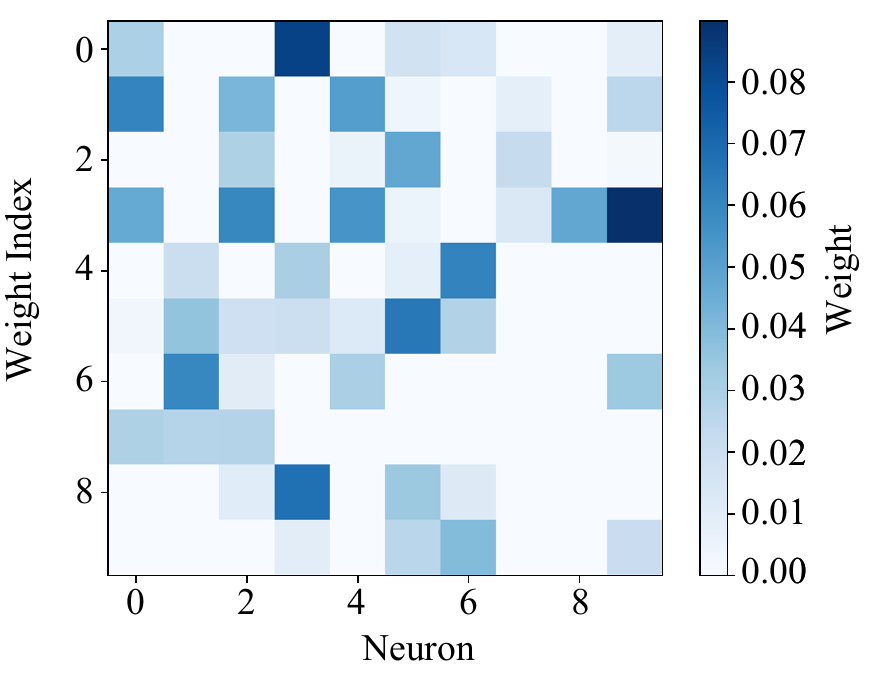}} 
	\vspace{0.5cm}
	\vspace{-5.5mm}
	\subfigure[The weights after pruning]{
		\includegraphics[width=7.5cm,height=5.7cm]{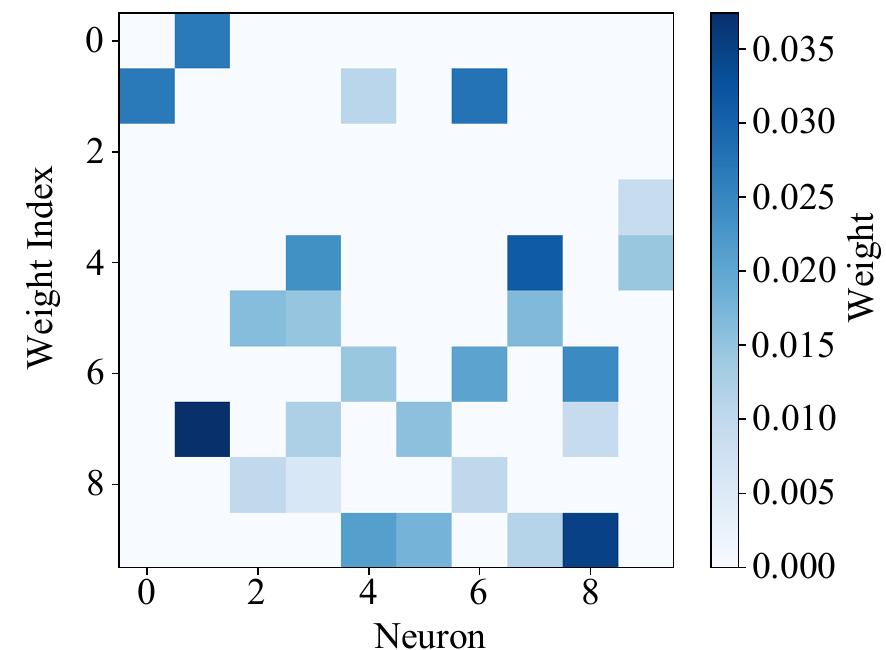}}
	\vspace{0.5cm}
	\vspace{-5.5mm}
	\caption{The impact of pruning on weights}
	\vspace{-0.5mm}
	\label{fig:15}
\end{figure}

\noindent
To visually illustrate the pruning results, a random layer of the network is selected, and the pruning of the first 10 weights of the first 10 neurons in that layer is visualized using the Matplotlib tool. The weight distribution before pruning is shown in Figure \ref{fig:15}(a), where white represents zero weights, and darker colors. The weights of this layer after pruning are depicted in Figure \ref{fig:15}(b), demonstrating a noticeable increase in sparsity in the weight distribution after model pruning. This increased sparsity allows the model to more flexibly adapt to different resource constraints in practical applications.
\par
\begin{figure}[htb]
	\centering
	\vspace {-3.5mm}
	\includegraphics[width=\linewidth]{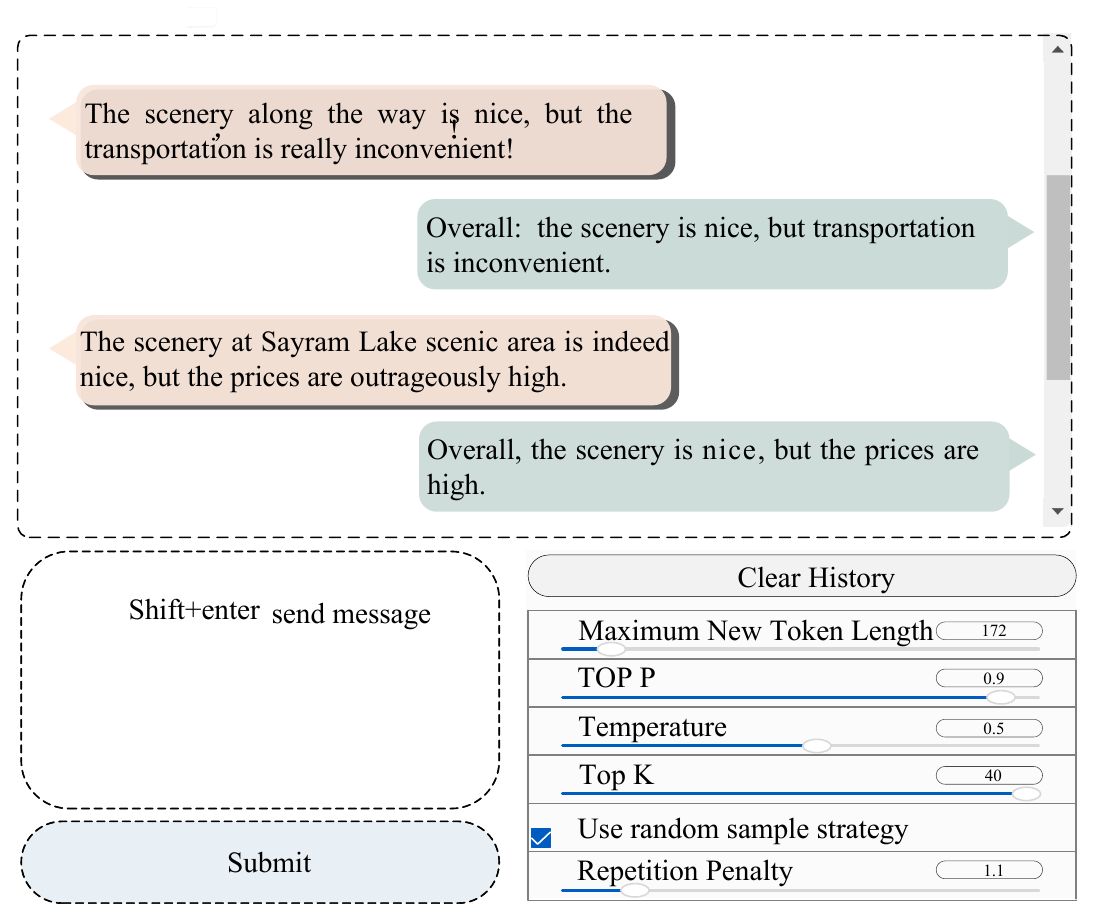}
	\vspace {-6.5mm}
	\caption{Diagram of auxiliary knowledge generation}
	\label{fig:16}
\end{figure}

Based on the pruned Atom-7B model, we demonstrated the auxiliary knowledge generation stage using Gradio, as shown in Figure \ref{fig:16}. In the input area in Figure \ref{fig:16}, users can input tourism reviews to generate corresponding auxiliary knowledge. The text generation process can be configured by adjusting parameters in the bottom right corner, where max\_new\_tokens determines the maximum number of new tokens generated, and repetition\_penalty serves as a penalty coefficient to prevent repetitive tokens.
\par
\begin{figure}[htb]
	\centering
	\vspace {-3.5mm}
	\includegraphics[width=\linewidth]{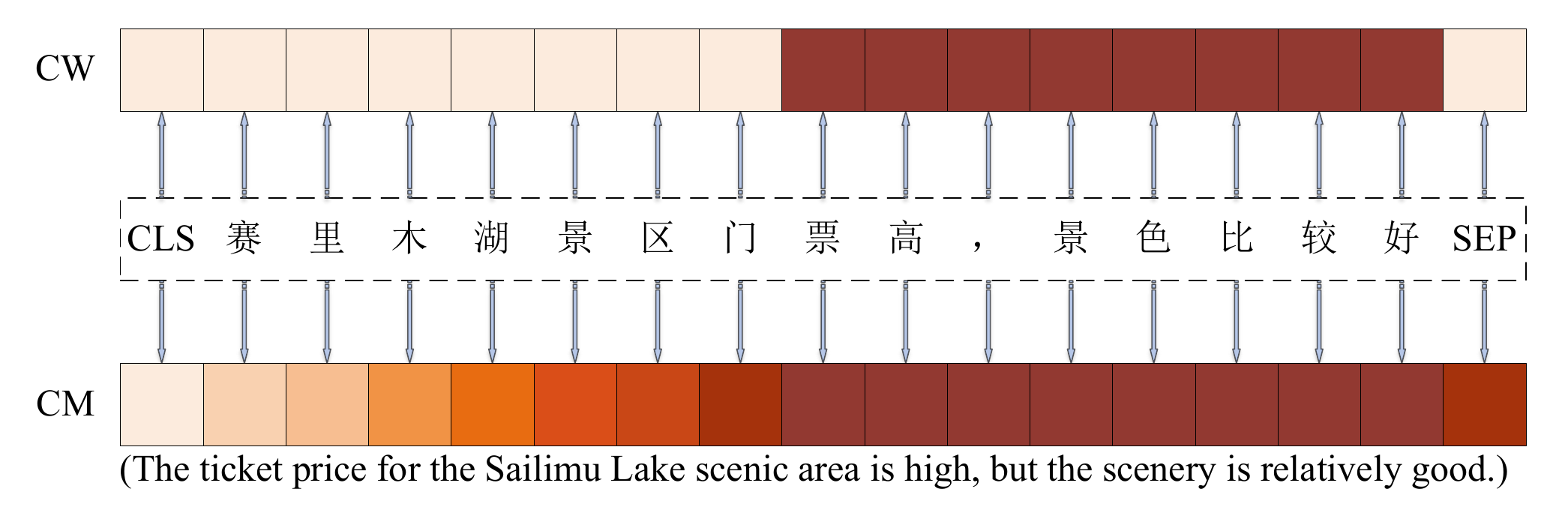}
	\vspace {-6.5mm}
	\caption{Visualization of distance information}
	\label{fig:17}
\end{figure}

cm[$i$] and cw[$i$] are used to represent the importance of each position in the text to the sentiment of the entity, helping to accurately capture contextual information around the entity. The corresponding weights are calculated based on the distance between the entity position and the current position in this method, indicating the direct impact of this position on the sentiment of the entity. For example, when allocate weights to the text for sentiment classification regarding the aspect ``scenery'', the outcome is depicted in Figure \ref{fig:17}. It can be observed that the weights of the aspect term ``scenery'' and the surrounding three characters are set to 1, and as the distance increases, their corresponding weights decrease. This method assists the model in distinguishing sentiment information at different positions in the text, thereby improving the accuracy of aspect-based sentiment analysis.
\par
\section{Conclusion}
\noindent
Traditional pipeline models often face challenges in handling ACOSQE tasks, including missing implicit sentiment elements, complex module integration, error propagation, and error accumulation. To tackle these challenges, we propose an aspect-based sentiment analysis model for tourism. Firstly, Adalora is used to fine-tune Atom-7B, generating high-quality auxiliary knowledge that enables the model to maintain its performance with limited training data. After training, Sparsegpt prunes the fine-tuned model to 50\% sparsity, lightening the model and thereby improving resource utilization efficiency. Then, the obtained auxiliary knowledge is combined with the original text and used as input for the ACOSQE module. Finally, distance information is utilized to assist in predicting the sentiment tendencies of different entities. Experiments conducted on a self-constructed tourism dataset and the public datasets Rest15 and Rest16 show improvements in F1 scores by 7.49\%, 0.05\%, and 1.06\%, respectively. This indicates improved performance in aspect-level sentiment analysis within the tourism sector and reasonable generalizability across other datasets. Additionally, the model's effective transferability to other aspect-based sentiment analysis tasks is confirmed through tests on CS, TASD, and ASTE tasks.
\section{Acknowledgment}
\noindent
This work was supported via funding from the National Natural Science Foundation of China (No.62266041).

\begin{biography}[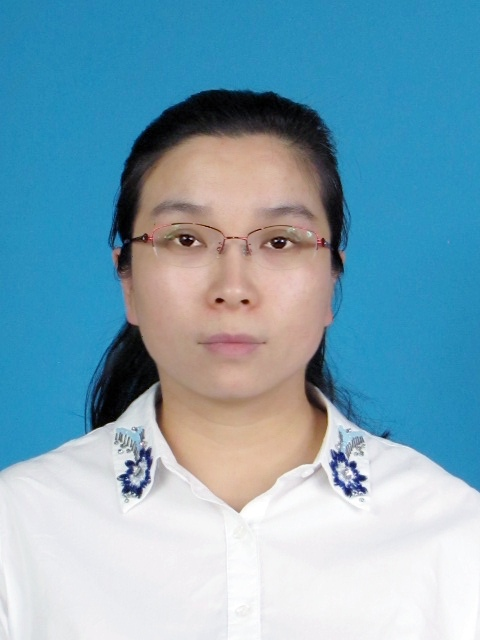]
	\noindent
	\textbf{Chun Xu} obtained the bachelor's degree in textile engineering from Xinjiang University, China in 2000, the master's degree in computer application technology from University of Chinese Academy of Sciences, China in 2005, the Ph.D. in computer application technology from the University of Chinese Academy of Sciences, China in 2018. Now, she is a professor at the college of Information Management of Xinjiang University of Finance and Economics, mainly engaged in research on travel data analysis and natural language processing.
\end{biography}
\begin{biography}[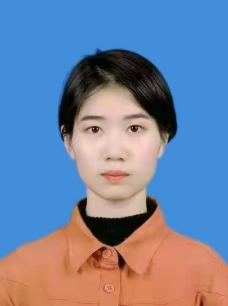]
	\noindent
	\textbf{Mengmeng Wang} obtained the bachelor of engineering degree from Chengdu Technological University, China in 2021. She is currently a master's student at the School of Information Management, Xinjiang University of Finance and Economics. Her primary research interest is travel data analysis and natural language processing.
\end{biography}
\begin{biography}[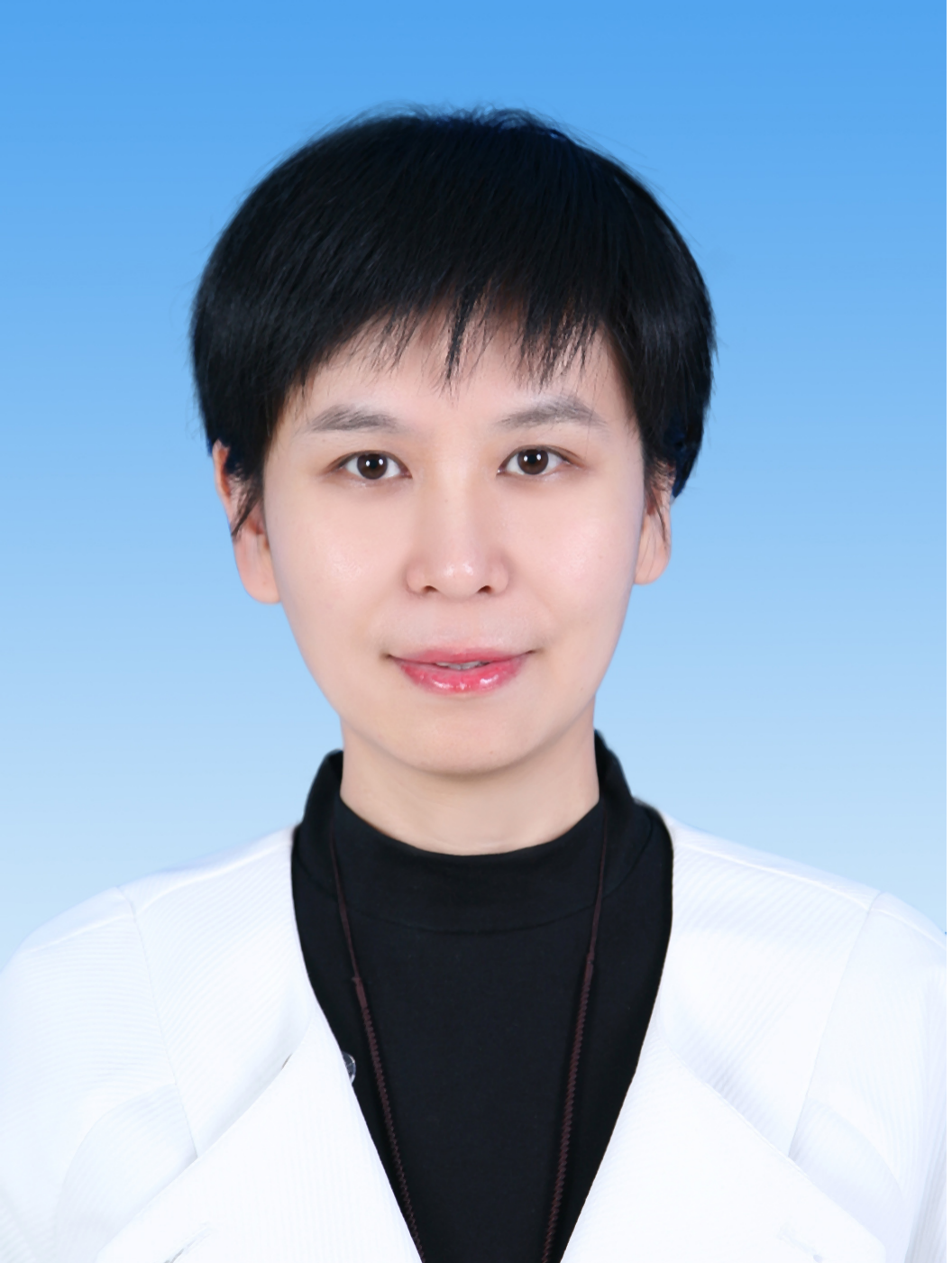]
	\noindent
	\textbf{Yan Ren} obtained the bachelor's degree from computer science and technology from Xinjiang Normal University, China in 2001, and the master's degree from computer technology from Dalian University of Technology, China in 2006. She serves as an associate professor at the School of Information Management of Xinjiang University of Finance and Economics, focusing on research in the areas of big data analysis and multimedia technology.
\end{biography}
\begin{biography}[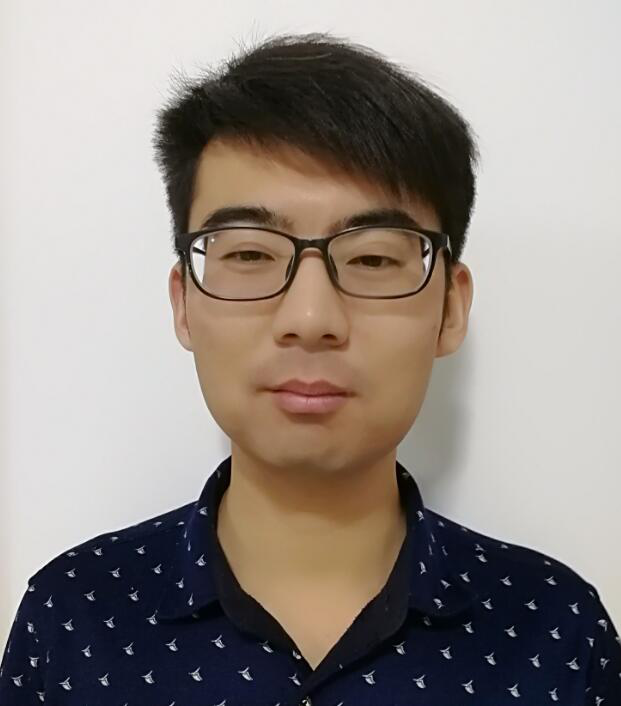]
	\noindent
	\textbf{Shaolin Zhu} obtained the the bachelor in engineering degree from Shenyang Aerospace University, China in 2013, the Ph.D. in computer application technology from the University of Chinese Academy of Sciences, China in 2018. He is a research associate of Computer Science at Tianjin University. His research focuses on natural language processing, specifically machine translation, and natural language generation.
\end{biography}
\clearpage
  \end{document}